\documentclass{bmvc2k}

\usepackage{graphicx}
\usepackage{booktabs}
\usepackage{amsmath}
\usepackage{amssymb}
\usepackage{multirow}
\usepackage[dvipsnames]{xcolor}
\usepackage{orcidlink}
\usepackage{pifont}

\title{DF-MoE: Generalizable Deepfake Detection via Multimodal Sparse Mixture-of-Experts}

\addauthor{Vlad Hondru\vspace{-0.35cm}}{}{1}
\addauthor{Florinel Alin Croitoru\vspace{-0.35cm}}{}{1}
\addauthor{Iuliana Georgescu\vspace{-0.35cm}}{}{1}
\addauthor{A. Sophia Koepke\vspace{-0.35cm}}{}{2,3}
\addauthor{Radu Tudor Ionescu}{raducu.ionescu@gmail.com}{1}
\addinstitution{
University of Bucharest\\
Bucharest, Romania
}
\addinstitution{
Technical University Munich, MCML\\
Munich, Germany
}
\addinstitution{
University of T\"{u}bingen, T\"{u}bingen AI Center\\
T\"{u}bingen, Germany
}

\runninghead{Hondru et al.}{DF-MoE: Deepfake Detection via Mixture-of-Experts}

\def\ie{\emph{i.e}\bmvaOneDot}
\def\eg{\emph{e.g}\bmvaOneDot}

\def\etal{\emph{et al}\bmvaOneDot}

\newcommand{\xmark}{\ding{55}}

\definecolor{best}{RGB}{0,80,200}        
\definecolor{second}{RGB}{220,120,0}    

\begin{document}

\maketitle

\setlength{\abovedisplayskip}{3.8pt}
\setlength{\belowdisplayskip}{3.8pt}
\setlength{\abovedisplayshortskip}{3pt}
\setlength{\belowdisplayshortskip}{3pt}

\begin{abstract}
Audio-visual deepfake detection is an actively studied topic, where one of the main challenges is to develop detectors able to generalize across deepfake generation methods. We conjecture that overfitting can be mitigated by extracting multiple high-level cues from the available audio and visual modalities via pre-trained models. We therefore assemble a wide variety of pre-trained models to extract features that encode mouth movements, face parsing, facial expressions, head pose, gaze tracking, heart rate, audio emotion and speech activity. We further integrate both unimodal and multimodal cues via a Mixture-of-Experts (MoE) backbone to detect deepfakes. We perform in-domain and cross-domain experiments on five benchmarks for deepfake detection (MAVOS-DD, AVLips, PolyGlotFake, BioDeepAV, FakeAVCeleb) to compare our framework (DF-MoE) with state-of-the-art methods. Our results indicate that DF-MoE obtains superior deepfake detection results, surpassing all competing methods. We release our code at \url{https://github.com/vladhondru25/DF-MoE}.
\end{abstract}

\begin{figure}[h]
 \vspace{-0.2cm}
    \centering
    \includegraphics[width=0.9\linewidth]{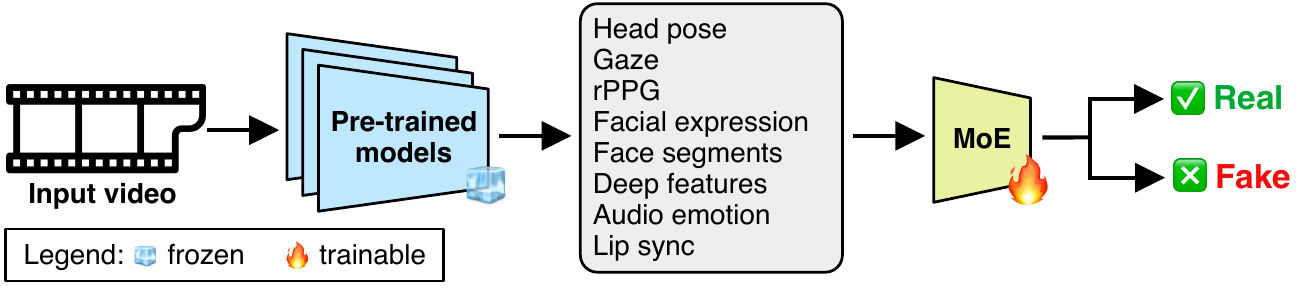}
    \vspace{-0.2cm}
    \caption{Overview of the proposed pipeline for deepfake detection. First, we employ multiple pre-trained models to extract high-level semantic cues. Then, a lightweight Mixture-of-Experts (MoE) transformer aggregates the extracted signals and learns to classify audio-visual inputs as real or fake. Our design prevents overfitting to a specific dataset by keeping the pre-trained models frozen, while only training the lightweight MoE on the target task.}
    \label{fig:tease}
    \vspace{-0.7cm}
\end{figure}

\section{Introduction}
\label{sec:intro}
\vspace{-0.1cm}

With the continuous advancements in generative AI \cite{Chen-AAAI-2025,Guo-ArXiv-2024,Ji-CVPR-2025,Wang-IJCAI-2021,Zheng-ArXiv-2024} and the increasing democratization of generative technology, concerns towards potential misuse are rising \cite{Croitoru-Arxiv-2024}. Deepfakes, \ie generated multimedia (audio-visual) content that aims to deceive humans into believing that the content is real, are predominantly used by malicious users in financial scams, misinformation and identity theft. Since humans, especially vulnerable groups (\eg elders, visually impaired, etc.), may be easily deceived by deepfakes, an important area of research is the development of deepfake detectors \cite{Cao-CVPR-2022,Hou-ICPR-2024,oorloff-CVPR-2024,Qian-ECCV-2020,Smeu-CVPR-2025,xu-ICCV-2023,Yan-CVPR-2025,Yan-ICCV-2023,Zou-ICASSP-2024}. One of the main challenges in this area is to obtain detectors able to generalize across deepfake generation methods, a key characteristic that is required to overcome the restless development of generative AI technology. Despite its importance, most studies in deepfake detection leave this aspect aside, focusing on improving performance across existing deepfake detection benchmarks \cite{Qian-ECCV-2020, Roessler-ICCV-2019, Cao-CVPR-2022, Huang-CVPR-2023, Shiohara-CVPR-2022}. While many works report impressive performance (up to $99\%$ accuracy) in deepfake detection on public benchmarks, they fail to determine the generalization capacity of the detectors on deepfakes produced by newer and more advanced generative methods. 
Against the mainstream practice in the field, several studies attempted to address the generalization concern revolving around deepfake detectors \cite{Nadimpalli-CVPRW-2022,Lai-ACMMM-2025, Ma-NeurIPS-2025,Yermakov-WACV-2026}. For instance, Ma \etal~\cite{Ma-NeurIPS-2025} noticed that current deepfake models create either face inconsistency or up-sampling artifacts, and leveraged this information to generate pseudo-fake training data.  
Lai \etal~\cite{Lai-ACMMM-2025} proposed the Generalized Multi-Scenario Deepfake Detection framework to enable jointly training a model on multiple datasets, since naively training a deepfake model on multiple deepfake datasets decreases the overall performance. 

Unlike previous studies in this area \cite{Nadimpalli-CVPRW-2022,Lai-ACMMM-2025, Ma-NeurIPS-2025,Yermakov-WACV-2026}, which usually fine-tune models on deepfake data, we propose to overcome the dataset overfitting issue by extracting multiple high-level cues from the available audio and video modalities via pre-trained models, as shown in Figure \ref{fig:tease}. Our assumption is that models that are pre-trained on distinct tasks, \eg facial expression recognition, audio emotion recognition, head pose estimation, or gaze tracking, provide meaningful information about the authenticity of audio-visual content, without risking overfitting to artifacts specific to a certain generative model. We therefore aggregate such pre-trained models into a unified pipeline, while deferring the training stage on real vs.~fake content classification to a lightweight Mixture-of-Experts backbone \cite{Shazeer-ICLR-2017} that combines all the high-level (semantic) features. To avoid learning the distinctive patterns of different deepfake generation methods, we further introduce a novel contractive-repulsive objective (CRO) that contracts latent vectors around corresponding class anchors (aiming to reduce distinctive patterns inside each class), while repelling class anchors beyond a given margin (aiming to enforce better discrimination between classes). By expressing our contrastive objective through class anchors, we avoid searching for positive/negative pairs via expensive hard sample mining \cite{Georgescu-MVA-2022,Georgescu-ICPR-2021,Harwood-ICCV-2017,Schroff-CVPR-2015,Suh-CVPR-2019}. Our \textbf{d}eep\textbf{f}ake detection based on sparse \textbf{M}ixture-\textbf{o}f-\textbf{E}xperts (DF-MoE) is not only designed to improve generalization capacity, but also to maintain a reasonable compute time, achieving near real-time processing speed on a single Nvidia RTX 5090 GPU with 32GB VRAM.
    
We perform experiments on five recent and challenging benchmarks (MAVOS-DD \cite{Croitoru-ArXiv-2025}, AVLips \cite{Liu-NeurIPS-2024}, PolyGlotFake \cite{Hou-ICPR-2024}, BioDeepAV \cite{Croitoru-Arxiv-2024}, FakeAVCeleb \cite{khalid-NeurIPS-2021}), showing that DF-MoE outperforms state-of-the-art deepfake detectors \cite{Afchar-WIFS-2018,Cao-CVPR-2022,Dang-CVPR-2020,Hou-ICPR-2024,khalid-NeurIPS-2021,Li-CVPRW-2018,Ni-CVPRW-2022,oorloff-CVPR-2024,Qian-ECCV-2020,Roessler-ICCV-2019,Smeu-CVPR-2025,tan-ICML-2019,xu-ICCV-2023,Yan-CVPR-2025,Yan-ICCV-2023,Yermakov-WACV-2026,Zou-ICASSP-2024} across all five datasets, in both open-set and cross-domain evaluation scenarios. Ablation studies confirm that standalone high-level features from pre-trained models are individually well-suited for deepfake detection. Yet, combining all features via MoE is the best way to unleash their potential towards generalizable deepfake detection. Finally, we show that our framework takes a leap forward towards interpretable decisions via attributing gradient weights to the high-level features from pre-trained models, \ie DF-MoE can determine that a video clip is fake because it exhibits unusual facial expressions, inconsistent speech and mouth movements, unexpected gaze behavior, etc.

In summary, our contribution is fourfold:
\begin{itemize}
    \item \vspace{-0.1cm} We propose DF-MoE, a novel deepfake detection framework that employs a sparse Mixture-of-Experts transformer to integrate multiple complementary cues into an end-to-end pipeline. 
    \item \vspace{-0.05cm} We introduce CRO, a novel contractive-repulsive (contrastive) loss that improves the generalization capacity across deepfake generation methods, harnessing learnable class anchors to improve latent space organization, while avoiding expensive hard sample mining techniques employed by conventional contrastive learning objectives.
    \item \vspace{-0.05cm} We demonstrate the high generalization capacity of DF-MoE in a suite of challenging open-set and cross-domain deepfake detection experiments, where DF-MoE surpasses state-of-the-art deepfake detectors.
    \item \vspace{-0.05cm} We show that DF-MoE provides a leap forward towards interpretable decisions, by inherently determining which high-level features contribute to the final decision. 
\end{itemize}

\vspace{-0.2cm}
\section{Related Work}
\vspace{-0.1cm}

Based on the input domain, deepfake detection methods can be categorized into image-level~\cite{Chen-CVPR-2022, Dong-CVPR-2023, Lin-CVPR-2024b, Yan-CVPR-2024, Tan-CVPR-2024, Nguyen-CVPR-2024, Yao-ICCV-2023, Yan-ICCV-2023, Chen-NeurIPS-2022, Tan-AAAI-2024, Le-AAAI-2024, Kim-CVPRW-2021, Xu-WACVW-2023, Du-CIKM-2019, Huang-CVPR-2023, Shiohara-CVPR-2022, Zhao-CVPR-2021, Dong-ECCV-2022, Le-ICCV-2023,Larue-ICCV-2023, Sun-ICCV-2023, Zhao-ICCV-2021, Hooda-WACV-2024, Ju-WACV-2024, Tantaru-WACV-2024, Trinh-WACV-2021, Ba-AAAI-2024, Yang-AAAI-2022, Nirkin-TPAMI-2022, Lanzino-CVPRW-2024, Ciamarra-WACVW-2024, Jeong-WACV-2022, Ricker-CVPR-2024, Choi-ArXiv-2024, Li-NeurIPS-2024}, audio-level~\cite{Zhang-ICLR-2025, Tak-ICASSP-2021, Wang-INTERSPEECH-2023, Hua-SPL-2021, Tak-INTERSPEECH-2021, Jung-ICASSP-2022, Dong-CVPR-2022, Wang-ICMR-2022, Aghasanli-ICCVW-2023} and multimodal methods~\cite{Raza-CVPR-2023, Cozzolino-CVPR-2023, Kihal-MTA-2023, zhou-ICCV-2021, oorloff-CVPR-2024, Salvi-JI-2023, Ilyas-ASC-2023, Asha-MS-2024, Liu-SPIC-2023, Feng-CVPR-2023, Zou-ICASSP-2024, Nie-ACMMM-2024, Zhang-ACM-2024, Smeu-CVPR-2025, Kong-NeurIPS-2025, Yan-CVPR-2025, Han-CVPR-2025, Astrid-ICASSP-2025}. The initial efforts have employed convolutional neural networks (CNNs)~\cite{Chen-CVPR-2022, Dong-CVPR-2023, Lin-CVPR-2024b, Yan-CVPR-2024, Tan-CVPR-2024, Nguyen-CVPR-2024, Yao-ICCV-2023, Yan-ICCV-2023, Chen-NeurIPS-2022, Tan-AAAI-2024, Le-AAAI-2024, Kim-CVPRW-2021, Xu-WACVW-2023, Du-CIKM-2019, Huang-CVPR-2023, Shiohara-CVPR-2022, Zhao-CVPR-2021, Dong-ECCV-2022, Le-ICCV-2023,Larue-ICCV-2023, Sun-ICCV-2023, Zhao-ICCV-2021, Hooda-WACV-2024, Ju-WACV-2024, Tantaru-WACV-2024, Trinh-WACV-2021, Ba-AAAI-2024, Yang-AAAI-2022, Nirkin-TPAMI-2022, Lanzino-CVPRW-2024, Ciamarra-WACVW-2024, Jeong-WACV-2022, Cheng-CVPR-2025, Sun-IJCV-2025, Ricker-CVPR-2024, Choi-ArXiv-2024, Li-NeurIPS-2024} and recurrent neural networks (RNNs)~\cite{guera-AVSS-2018, sabir-CVPR-2019, hu-AAAI-2022, Liu-WACV-2023, montserrat-CVPR-2020}. More recently, most studies have adopted the transformer architecture~\cite{Bartusiak-ACSSC-2021, zhou-ICCV-2021, oorloff-CVPR-2024, Salvi-JI-2023, Ilyas-ASC-2023, Asha-MS-2024, Liu-SPIC-2023, Feng-CVPR-2023, Zou-ICASSP-2024, Nie-ACMMM-2024, Zhang-ACM-2024, Smeu-CVPR-2025, Kong-NeurIPS-2025, Yan-CVPR-2025, Han-CVPR-2025}, with the primary goal of jointly analyzing the video and audio content.  
Consequently, the recently proposed state-of-the-art methods~\cite{oorloff-CVPR-2024,Zou-ICASSP-2024,Astrid-ICASSP-2025, zhou-ICCV-2021} for deepfake detection are multimodal.

Zhou~\etal~\cite{zhou-ICCV-2021} proposed one of the early frameworks on jointly modeling  audio and video streams to perform deepfake detection. The framework was based on the synchronization between video and audio, and different fusion approaches, such as late-fusion and two-plus-one. 
Several other works~\cite{Astrid-ICASSP-2025,oorloff-CVPR-2024} also relied on the synchronization between audio and video. For instance, Astrid~\etal~\cite{Astrid-ICASSP-2025} employed pseudo-fake generation to improve the detection of local inconsistencies between audio and video. 
AVFF~\cite{oorloff-CVPR-2024} captures the correspondence between audio and video modalities by first pre-training the model on real training data, then performing supervised training using both unimodal and cross-modal features. 
Switching to a more fine-grained output, Delocate~\cite{Hu-IJCAI-2024} performs both detection and localization of deepfakes in videos in two stages, by first reconstructing the faces in the frames, and then classifying them. 
For a more efficient training, Hui~\etal~\cite{Hui-AAAI-2025} proposed multi-task audio-visual prompt learning, by injecting prompts into each layer of an audio-visual foundation model, without updating the entire model. To efficiently process a video, TALL~\cite{xu-ICCV-2023} transforms it into a thumbnail, and then applies the image-level Swin Transformer \cite{Liu-ICCV-2021} to perform video deepfake detection. 

Several works~\cite{Nadimpalli-CVPRW-2022,Lai-ACMMM-2025, Ma-NeurIPS-2025,Yermakov-WACV-2026} focused on the generalization of the deepfake detectors to out-of-domain data. Ma~\etal~\cite{Ma-NeurIPS-2025} categorized the deepfake artifacts into face inconsistency and up-sampling artifacts, noticing that existing deepfake models exhibit either or both kinds of artifacts. Leveraging this observation, Ma~\etal~\cite{Ma-NeurIPS-2025} generated pseudo-fake data to train their deepfake model, obtaining a robust classifier.
Nadimpalli~\etal~\cite{Nadimpalli-CVPRW-2022} tackled the deepfake detection task using a hybrid combination of supervised and reinforcement learning techniques to improve cross-domain performance. 
Lai~\etal~\cite{Lai-ACMMM-2025} observed that naively training a model on multiple datasets does not improve the joint performance for deepfake detection. Therefore, they proposed the Generalized Multi-Scenario Deepfake Detection framework to enable the joint training of a model on multiple datasets, including a domain-aware meta-learning strategy.
Closer to our work, optimizing for  both increased generalization and short training time, 
Yermakov~\etal~\cite{Yermakov-WACV-2026} fine-tuned only the layer normalization parameters of a foundational pre-trained vision encoder. They obtained good results on several cross-dataset scenarios.

Different from previous works on generalizable deepfake detection~\cite{Nadimpalli-CVPRW-2022,Lai-ACMMM-2025, Ma-NeurIPS-2025,Yermakov-WACV-2026}, 
we propose to extract different signals from available pre-trained models, overcoming overfitting and increasing the out-of-domain generalization capabilities by keeping these models frozen. To our knowledge, we are the first to integrate a wide range of high-level semantic cues for deepfake detection. The integration is performed by a learnable MoE block, which is carefully trained to mitigate overfitting on deepfake generative models seen at training via a novel contractive-repulsive objective.


\vspace{-0.2cm}
\section{Method}
\vspace{-0.1cm}

\begin{figure*}[t]
    \begin{center}
    \includegraphics[width=\linewidth]{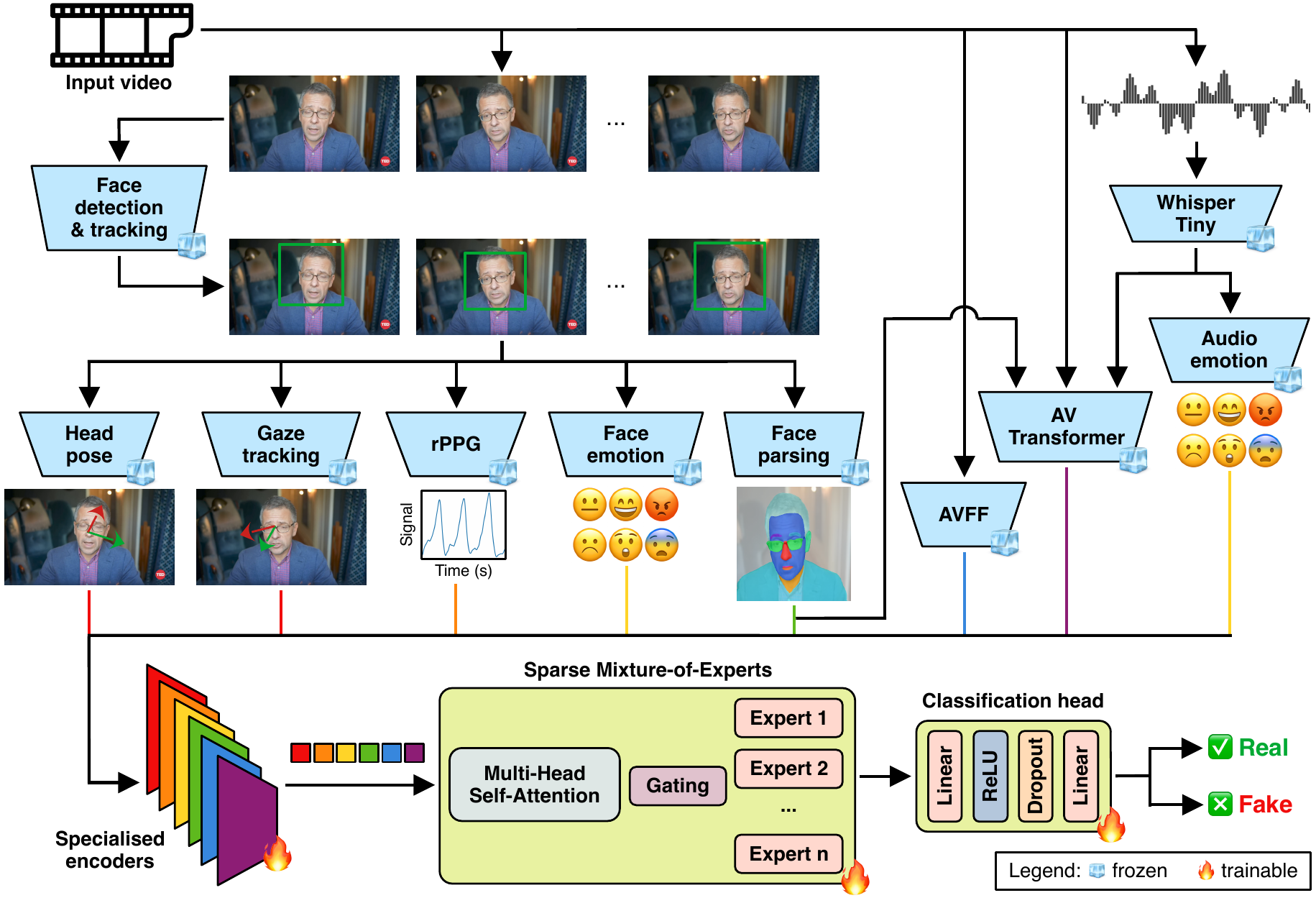}
    \end{center}
    \vspace{-0.55cm}
    \caption{Detailed overview of our DF-MoE framework for deepfake detection. Face detection and tracking are applied to obtain face tracks. Next, multiple pre-trained models (frozen) are employed to extract high-level semantic cues (features). The extracted features are encoded into tokens via specialized encoders (trainable adapters), ensuring consistent dimensionality across different cues. The resulting tokens are processed by a sparse MoE transformer (trainable). Best viewed in color.}
    \label{fig:pipeline}
    \vspace{-0.4cm}
\end{figure*}

Spotting a deepfake video requires reasoning over multiple signals simultaneously, especially for videos with photorealistic facial manipulations generated by state-of-the-art models, in which the artifacts are very difficult to recognize, even for humans. Nevertheless, subtle inconsistencies often persist across time: unnatural head movements, irregular gaze patterns or audio-visual mismatches. To this end, we introduce DF-MoE, a multimodal deepfake detection framework that harnesses high-level semantic cues extracted from pre-trained models. We aggregate the semantic cues and model their temporal dynamics via a sparse MoE architecture. Instead of directly operating on raw pixels, our design decomposes the task into (\emph{i}) extracting semantically meaningful visual and audio descriptors, and (\emph{ii}) learning cross-modal and temporal inconsistencies to capture the manipulated content. We showcase our full pipeline in Figure~\ref{fig:pipeline}, and explain its components in detail below.

\vspace{-0.2cm}
\subsection{Extracting Visual and Audio Cues}
\vspace{-0.1cm}

Our pipeline for feature extraction begins by splitting each video into frames. Then, on every fifth frame, we apply a face detector based on YOLOv11 \cite{khanam-arxiv-2024} to extract the faces. The following step is to employ a face tracking model. For each pair of consecutive processed frames, we employ a face tracker based on DeepSORT~\cite{wojke-ICIP-2017} to match the faces and form long-term continuous tracking paths along the video. Each tracking path in a video corresponds to one person, and is processed independently through the pipeline. 
For each face track, we apply pre-trained models to extract a series of visual features: head pose, eye gaze, face segmentation, remote photoplethysmography (rPPG), and facial expression. At the same time, we take the associated audio signal and extract speech embeddings with Whisper-Tiny \cite{radford-whisper-2022}. These are also used to compute emotion features from a custom pre-trained model for audio emotion recognition. 

\noindent
\textbf{Head pose.} Deepfake methods may produce head movements with unnatural rigidity, subtle temporal jitter or physically implausible transitions (especially portrait animation models), which can be detected by modeling their temporal consistency. Therefore, for each frame, we estimate the 3D head pose using HopeNet~\cite{Ruiz-CVPR-2018}, which returns three head pose angles (roll, pitch and yaw), denoted as $F^3_{\text{HP}}$. 

\noindent
\textbf{Gaze estimation.} Synthetic videos often fail to preserve natural gaze stability and coordination between both eyes, leading to measurable irregularities over time. The focus of the eyes on a specific point (gaze) is estimated by regressing the three Euler angles with a ResNet-34~\cite{He-CVPR-2015}. We drop the roll angle, keeping only the horizontal and vertical angles, as the eyes cannot rotate along the roll direction. For each frame, its gaze features are denoted as $F^2_{\text{Gaze}}$. 

\noindent
\textbf{Face segmentation.} We hypothesize that analyzing the temporal consistency of face parts can capture micro-movements and subtle spatial inconsistencies. To capture inconsistencies, we apply the Bilateral Segmentation Network~\cite{yu-ECCV-2018} and obtain a facial segmentation map $F^{512 \times 512}_{\text{Seg}}$ with labeled face parts.

\noindent
\textbf{rPPG.} Since heart rate is estimated via a weak rPPG signal (mostly invisible to humans), we conjecture that deepfake video generators do not insert such weak signals in generated video. Therefore, extracting rPPG signals can help distinguish between real and fake videos. We extract remote photoplethysmography $F^{5 \times 512}_{\text{rPPG}}$ via the model proposed by Yue \etal~\cite{yue-IJCV-2025}. This model captures subtle periodic color variations in facial skin regions caused by blood flow. The rPPG signal can provide an estimate of the subject's physiological pulse (heart rate) without physical contact. We estimate the rPPG waveform for five different face points. 


\noindent
\textbf{Facial expression.} Deepfakes can be characterized by irregular temporal transitions between facial expression classes or by inconsistent facial expression and speech emotion classes. Therefore, given only the visual input, a facial expression recognition (FER) model is applied to classify the face in each frame into one of the eight basic emotions: anger, contempt, disgust, fear, happiness, neutral, sadness and surprise. We employ an EfficientNet-B0~\cite{tan-ICML-2019} pre-trained on Video-level Group Affect~\cite{Sharma-ACIIW-2019}, which produces a vector of probabilities across the eight emotion classes, denoted as $F^8_{\text{FER}}$. 

\noindent
\textbf{Speech embeddings.} Since speech signal is imperative for audio-visual deepfake detection, we take a pre-trained Whisper-Tiny \cite{radford-whisper-2022} and encode speech via its latent representation $F^{384}_{\text{Speech}}$. The speech embeddings contain information about the phonetic content, capturing potential irregularities of synthesized or voice-cloned speech. 

\noindent
\textbf{Audio emotion.} We aim to complement facial expressions with emotion classes from the audio input. To this end, we pre-train a simple MLP classifier over the extracted speech embeddings 
for speech emotion recognition (SER) on the CREMA-D \cite{Houwei-TAC-2014} dataset. The resulting output, denoted as $F^8_{\text{SER}}$, provides cues about the temporal consistency of speech emotion, that can be used for assessing the consistency between audio and visual emotion. 

\noindent
\textbf{Lip sync.} Lip syncing is a crucial signal for detecting deepfakes. Therefore, we pre-train a custom transformer that aims to detect inconsistencies between the lip movement and speech. We use the output of the face parser (Bilateral Segmentation Network~\cite{yu-ECCV-2018}) to obtain a crop around the mouth. We concatenate $50$ crops 
and obtain their DINOv3 \cite{simeoni-arxiv-2025} embeddings. Similarly, we extract the Whisper-Tiny embeddings from the corresponding audio. The resulting visual and audio embeddings are fed into our cross-modal audio-visual (AV) transformer. The model consists of self-attention layers, followed by cross-attention layers, and outputs a 512-dimensional vector $F^{512}_{\text{LS}}$. 

\noindent
\textbf{AVFF.} AVFF is a two-stage transformer framework for deepfake detection \cite{oorloff-CVPR-2024}. The first stage adopts unimodal encoders and decoders along with cross-modal networks that are trained with reconstruction and contrastive objectives to capture audio-visual correspondences. The second stage fine-tunes the model on deepfake classification, so we discard this stage to prevent overfitting. 
Specifically, we take the unimodal audio and video representations before the classification stage. Each unimodal representation is a 1024-dimensional vector. We concatenate the two vectors to obtain a deep feature vector denoted as $F^{2048}_{\text{AVFF}}$. To mitigate overfitting on deepfake detection, we start from the pre-trained AVFF based on self-supervision and optimize only the unimodal transformer blocks via Effort \cite{Yan-ICML-2025}. This is a LoRA-style \cite{Hu-ICLR-2022} approach that decomposes the weight matrices via Singular Value Decomposition (SVD), freezes the principal components, and fine-tunes only the remaining components. Effort is specifically designed to offer strong generalization in deepfake detection \cite{Yan-ICML-2025}. Moreover, we keep the cross-modality fusion modules (A2V Network and V2A Network) frozen, since interactions between the audio and visual modalities can also contribute to overfitting.





\vspace{-0.2cm}
\subsection{Specialized Adapters}
\label{sec_spec_enc}
\vspace{-0.1cm}

To aggregate the high-level cues into a joint architecture, all features are projected into a shared embedding space of size $h = 128$, using specialized adapters.

\noindent
\textbf{Movement adapter.} The head pose $F^3_\text{HP}$ and eye gaze $F^2_\text{Gaze}$ are fed into a two-layer bidirectional LSTM \cite{Hochreiter-NC-1997}. Then, a cross-attention layer lets the gaze attend to the pose (gaze as queries, head pose as keys and values). In the end, the cross-attention output and the gaze representation (passed via a skip connection) are mean-pooled over time, concatenated, and projected. We denote the final head movement and gaze features as $E_\text{HP+Gaze} \in \mathbb{R}^{h}$.

\noindent
\textbf{Face parsing adapter.} The input of this encoder consists of the face segmentation maps $F^{512 \times 512}_\text{Seg}$. The architecture is represented by 3D CNN with 3D convolution and max pooling layers along time and space, followed by a global average pooling and linear layers. The resulting features are denoted as $E_\text{Seg}  \in \mathbb{R}^{h}$.

\noindent
\textbf{rPPG adapter.} 
The rPPG encoder takes the rPPG signal $F^{5 \times 512}_\text{rPPG}$ as input. It is composed of two 1D convolutional layers, followed by average pooling and two bidirectional GRUs. The final features are denoted by $E_\text{rPPG}  \in \mathbb{R}^{h}$.

\noindent
\textbf{Emotion adapter.} This encoder uses both discrete emotions (from audio $F^8_\text{AER}$ and video $F^8_\text{FER}$). Each stream is passed through a shared embedding layer, and then processed by separate bidirectional GRUs \cite{Cho-EMNLP-2014}. The resulting representations are combined using a cross-attention layer, concatenated with the video embeddings, and jointly projected. The resulting features are denoted by $E_\text{Emo}  \in \mathbb{R}^{h}$.

\noindent
\textbf{Lip sync adapter.} This encoder takes the representation from our custom AV transformer, pools it over time with an adaptive average and projects it into the shared embedding space. We denote the final lip sync features by $E_\text{LS} \in \mathbb{R}^{h}$.

\noindent
\textbf{Audio-visual adapter.} The features obtained with AVFF, $F^{2048}_{\text{AVFF}}$, are simply projected through a linear layer to the shared embedding space. We denote the resulting feature vector by $E_\text{AVFF} \in \mathbb{R}^{h}$.

\vspace{-0.2cm}
\subsection{Sparse Mixture-of-Experts}
\vspace{-0.1cm}

To classify a video as real or deepfake using the projected embeddings, we employ a sparse Mixture-of-Experts transformer. The projected embeddings are concatenated into a sequence $S$, as follows:
\begin{equation}
S =  \left[E_\text{HP+Gaze}, E_\text{Seg}, E_\text{Emo}, E_\text{rPPG}, E_\text{LS}, E_\text{AVFF}\right] \in \mathbb{R}^{6\times h}.
\end{equation}
The sequence $S$ is then processed by a multi-head self-attention module, enabling cross-feature modeling, before getting routed to the experts. The fact that experts can focus on different types of signals is particularly useful in our case, since tokens embed different high-level information from a wide variety of pre-trained models. 

\noindent
\textbf{Mixture-of-experts.} The gating network is represented by a linear layer, denoted as $g(\cdot)$, that gives a ranking score for the available experts. The gating mechanism assigns routing weights to indicate which expert is most suitable for a given token. Formally, for each token $s_i$ of the sequence $S=\{s_i\}_{i=1}^6$, we rank all experts based on the returned values $g(s_i)_j, \forall j\in \{1, \dots, n\}$, where $n$ is the number of experts. 
Subsequently, each token $s_i$ is passed through the top-$k$ scoring experts. Let $J_i = \{j_1^i, \dots, j_k^i\}, \forall i \in \{1,\dots,6\}$ denote the set of indices to which the token $s_i$ is routed.
Based on the ranking scores, we compute the weights $p_i$ that are used to determine the output tokens of the MoE layer, given the input tokens $S$. Specifically, we employ the following equation, obtaining normalized weights for the experts:
\begin{equation}
    p_{ij} = \frac{\exp({g(s_i)_j})}{\sum_{l\in J^i}{\exp({g(s_i)_l})}}, j\in J^i, i \in \{1, \dots, 6\}.
\end{equation}
After computing $\{p_i\}_{i=1}^6$, we can determine the output sequence $\hat{S}$. Formally, if we denote the experts with the top-$k$ highest scores in $p_i$ by $\{\text{e}_j(\cdot)\}_{j\in J^i}$, $\forall i \in \{1,\dots, 6\}$, then each output token $\hat{s}_i$ of the MoE layer is computed as:
\begin{equation}
    \label{eq:final_s}
    \hat{s}_i = \sum_{j=1}^{k}p_{ij} \cdot \text{e}_j(s_i).
\end{equation}
Intuitively, Eq.~\eqref{eq:final_s} uses the values in $p_i$ as weights for the representations returned by the experts. If an expert has a higher score in $p_i$, then its representation is more important in Eq.~\eqref{eq:final_s}.

\noindent
\textbf{Dropout regularization.}
At training time, we include a dropout regularization on the experts routing to avoid their over-specialization on a certain type of feature. The dropout is implemented by creating a mask $\{m_i\}_{i=1}^6 \in \{0, 1\}^n$ that corresponds to each score vector $\{g(s_i)\}_{i=1}^6$. We randomly set some of the positions in $m_i$ to 0, and the remaining ones to 1. The number of values assigned to 0 is controlled by a dropout rate $d=0.2$. The output of the dropout layer is given by $\hat{g}(s_i) = g(s_i) * m_i$,
where $*$ is the element-wise product. $\hat{g}(s_i)$ is further used in the top-$k$ routing logic described before.

\noindent
\textbf{Training losses.} Our final training objective comprises three components. The first one is the classic binary cross-entropy (BCE) loss over real and deepfake class labels. For this loss, we average the representations stored in the sequence $\hat{S}=\{\hat{s}_i\}_{i=1}^6$ to obtain the feature vector $R=\frac{1}{6} \sum_{\hat{s}_i \in \hat{S}} \hat{s}_i$ that is fed into the final classifier $c(\cdot)$. Given the ground-truth label $y \in \{0, 1\}$ of the input video, the binary cross-entropy loss is:
\begin{equation}
\label{eq:loss_cls}
    \mathcal{L}_{\text{BCE}} = - \left[y \cdot \log(c(R)) + (1 - y) \cdot \log(1 - c(R)) \right].
\end{equation}

If deepfakes are spread around in the latent space, \eg all around the real class region, it is likely that deepfakes generated by unknown generative methods will be confused with real samples. We therefore propose an additional objective to assist the standard BCE in structuring the latent space of the model, so as to reduce the discrimination power among various deepfake types, while boosting discrimination between real and fake samples. Our novel contractive-repulsive objective (CRO) is designed to reduce representation diversity inside classes (intra-class contraction), while increasing the gap between real and deepfake classes in the latent space (inter-class repulsion). This is achieved via a combination of three loss terms that operate with learnable class anchors. Let $A \in \mathbb{R}^h$ and $B \in \mathbb{R}^h$ denote the learnable anchors for the real and fake classes, respectively. We define the components of the CRO loss as follows:
\begin{equation}
\begin{split}
    \mathcal{L}_{\text{contract}}(A,B, R) &= (1-y)\cdot\|R - A \|_2^2 + y\cdot \|R-B \|_2^2,   \\
    \mathcal{L}_{\text{repel}}(A, B) &= \frac{1}{2}\left[ \max \left( 0, 2 \cdot M - \|A - B \| \right) \right]^{2}, \\
    \mathcal{L}_{\text{no-collapse}}(A, B)& = \frac{1}{2}  \left[ \max \left( 0, P - \| A \| \right) + \max \left( 0, P - \| B \| \right) \right]^{2},
\end{split}
\label{eq:cro_terms}
\end{equation}
where $M>0$ represents the margin between the two anchors, and $P$ is the minimum norm for each anchor. The anchors $A$ and $B$ are randomly initialized before training, and updated at every training iteration with the other trainable parameters. In our experiments, we set $M=1$ and $P=1$. $\mathcal{L}_{\text{contract}}$ minimizes the distance between the feature vector $R$ and its corresponding anchor, $A$ or $B$, depending on label $y$. $\mathcal{L}_{\text{repel}}$ enforces a minimum margin between the anchors $A$ and $B$, ensuring that classes are sufficiently far apart. Since class anchors $A$ and $B$ are learnable, a concentric configuration of real and fake latent vectors (\ie a cluster of real samples surrounded by a band of fake samples) might bring the anchors in the same vicinity, which might put the first two objectives into conflict. To avoid the collapse of class anchors, we introduce $\mathcal{L}_{\text{no-collapse}}$, which pushes anchors away from the origin. Finally, $\mathcal{L}_\text{CRO}$ is defined as:
\begin{equation}
\begin{split}
    \mathcal{L}_\text{CRO} &= \mathcal{L}_{\text{contract}}+\mathcal{L}_{\text{repel}}+\mathcal{L}_{\text{no-collapse}}. 
\end{split}
\label{eq:cro_loss}
\end{equation}
An advantage of our CRO loss over traditional contrastive losses is that it operates relative to a set of learnable anchors. Consequently, it achieves feature separation without requiring computationally-expensive procedures to mine adequate sample pairs \cite{Georgescu-MVA-2022,Georgescu-ICPR-2021,Harwood-ICCV-2017,Schroff-CVPR-2015,Suh-CVPR-2019}.


The third loss term penalizes the gating network if one expert receives too many tokens and the predicted probability for an expert is too high. This loss promotes diversity in the top-$k$ selected tokens. To compute this loss, we determine the fraction of tokens routed to each expert $\{f_j\}_{j=1}^n$ and the average probability score $\{\bar{p}_j\}_{j=1}^n$ assigned to each expert, where $\bar{p}_j = \frac{1}{6}\sum_{i=1}^6 p_{ij}$. 
In practice, both $f_j$ and $\bar{p}_j$ are estimated on an entire mini-batch.
These two variables are combined into a loss function defined as follows:
\begin{equation}
\label{eq:loss_aux}
    \mathcal{L}_{\text{aux}} = n \sum_{j=1}^{n} f_j \cdot \bar{p}_j.
\end{equation}
The final loss is a combination of those defined in Eq.~\eqref{eq:loss_cls}, Eq.~\eqref{eq:cro_loss} and Eq.~\eqref{eq:loss_aux}:
\begin{equation}
    \mathcal{L} = \mathcal{L}_{\text{BCE}} + \lambda_1 \cdot \mathcal{L}_{\text{CRO}} + \lambda_2\cdot\mathcal{L}_{\text{aux}},
\end{equation}
where $\lambda_1$ and $\lambda_2$ are hyperparameters that control the importance of the additional losses.

\vspace{-0.2cm}
\section{Experiments}
\vspace{-0.1cm}

\noindent
\textbf{Datasets.} To assess the generalization capacity of DF-MoE, we evaluate it on five datasets: AVLips \cite{Liu-NeurIPS-2024}, MAVOS-DD \cite{Croitoru-ArXiv-2025}, PolyGlotFake \cite{Hou-ICPR-2024}, BioDeepAV \cite{Croitoru-Arxiv-2024} and FakeAVCeleb \cite{khalid-NeurIPS-2021}. In terms of evaluation setups, we conduct (i) in-domain experiments on AVLips \cite{Liu-NeurIPS-2024} and MAVOS-DD \cite{Croitoru-ArXiv-2025}, (ii) open-set experiments on MAVOS-DD \cite{Croitoru-ArXiv-2025}, as well as (iii) cross-dataset experiments on PolyGlotFake \cite{Hou-ICPR-2024}, BioDeepAV \cite{Croitoru-Arxiv-2024} and FakeAVCeleb \cite{khalid-NeurIPS-2021}.

AVLips~\cite{Liu-NeurIPS-2024} is designed for training and evaluating lip-sync forgery detection models. The dataset contains over 3,000 real and 4,000 manipulated videos. Throughout our experiments, we utilize the official train and test splits. MAVOS-DD is a multilingual benchmark that provides a training set (21K videos), a validation set (4K videos) and four test sets (together containing over 60K test videos). The evaluation protocol is designed to support four setups: closed-set, open-set model, open-set language and open-set full. The \emph{closed-set} test set evaluates detectors on samples that correspond to the set of languages and generative methods seen during training. The \emph{open-set model} test set introduces new (unseen) generative methods. 
The \emph{open-set language} introduces two unseen languages, Hindi and German. 
Finally, the \emph{open-set full} test set simultaneously evaluates performance on unseen languages and unseen methods.








For the cross-dataset evaluation, we choose three recent datasets, PolyGlotFake~\cite{Hou-ICPR-2024}, BioDeepAV~\cite{Croitoru-Arxiv-2024} and FakeAVCeleb~\cite{khalid-NeurIPS-2021}. PolyGlotFake is a multi-lingual dataset that contains 766 real and 14,472 fake videos. Compared to MAVOS-DD, PolyGlotFake has two new languages, French and Japanese. The manipulation methods of PolyGlotFake consist of text-to-speech, voice conversion and lip-synchronization methods. BioDeepAV comprises 2,010 real and 1,693 fake videos. This dataset comprises diverse generative approaches, including NeRF-based synthesis, Gaussian Splatting, and diffusion models.  FakeAVCeleb is a very imbalanced dataset, comprising only 500 real and over 19,000 fake videos. We underline that typically employed metrics, such as AUC, can be misleading for highly imbalanced datasets. For example, if a model successfully isolates high-confidence fakes at the strictest thresholds, it rapidly increases the True Positive Rate (TPR), securing a high overall AUC. To ensure a correct evaluation and to avoid any bias in our evaluation metrics, we supplement the set of real videos with additional 19,000 videos randomly sampled from VoxCeleb2~\cite{chung-Interspeech-2018}. We highlight that this video addition is also suggested by the authors of FakeAVCeleb~\cite{khalid-NeurIPS-2021} in their original work. The resulting dataset is further referred to as Vox+FakeAVCeleb.


\noindent
\textbf{Baselines.}  We evaluate our method on AVLips and MAVOS-DD against several state-of-the-art methods: TALL \cite{xu-ICCV-2023}, MRDF~\cite{Zou-ICASSP-2024}, AVFF~\cite{oorloff-CVPR-2024}, AVH-Align \cite{Smeu-CVPR-2025}, AVH-Align$_\text{sup}$ \cite{Smeu-CVPR-2025}, RealForensics~\cite{haliassos-CVPR-2022}, LipForensics, and LipFD~\cite{Liu-NeurIPS-2024}. While most of these baselines leverage multimodal audio-visual features, TALL relies exclusively on video artifacts. For the cross-dataset evaluation on PolyGlotFake, BioDeepAV and Vox+FakeAVCeleb, we include additional baselines, \eg~UCF~\cite{Yan-ICCV-2023}, StA~\cite{Yan-CVPR-2025}, RECCE~\cite{Cao-CVPR-2022}, GenD \cite{Yermakov-WACV-2026}, among many others \cite{Afchar-WIFS-2018,Dang-CVPR-2020,Hou-ICPR-2024,khalid-NeurIPS-2021,Li-CVPRW-2018,Ni-CVPRW-2022,Qian-ECCV-2020,Roessler-ICCV-2019,Smeu-CVPR-2025,tan-ICML-2019,xu-ICCV-2023}. 

\noindent
\textbf{Ablated models.} We carry out ablation studies to assess the impact of each high-level cue integrated in DF-MoE. We also ablate the joint MoE module, employing a custom linear classifier instead, resulting in a version called DF-Linear. DF-Linear aggregates the same semantic cues as DF-MoE, serving as a critical ablation to isolate the performance gains brought by our MoE architecture. 

\noindent
\textbf{Hyperparameters.} All the models, including the baselines, are trained for 10 epochs, with the optimal checkpoint selected based on validation performance. For the baselines, the hyperparameters (learning rate, optimizer, etc.) are configured in accordance with the official recommendation from their corresponding publications. For DF-MoE and DF-Linear, we employ AdamW as the optimizer, with a learning rate of $10^{-4}$, and a batch size of $4$. The projection dimension for the shared latent space of the specialized encoders is set to $h=128$, the number of experts $n$ is set to $6$, and each token is routed to $k=2$ experts, based on the scores provided by the gating network. The specialized encoders vary in architecture, as per Section \ref{sec_spec_enc}. However, these encoders generally integrate a hidden bottleneck layer, with a latent representation of $64$ dimensions. The weight for the CRO loss is $\lambda_1=1$, and the weight of the auxiliary loss is set to $\lambda_2=0.01$. More reproducibility details are provided in the supplementary.

\begin{table}[t!]
  \centering
  \setlength\tabcolsep{0.1cm}
  \fontsize{8}{9}\selectfont{
  \begin{tabular}{lccc c ccc c ccc c ccc c ccc}
    \toprule
      \multirow{2.5}{*}{Method} &  \multicolumn{3}{c}{\multirow{1}{*}{Closed-set}} & & \multicolumn{3}{c}{\multirow{1}{*}{Open-set model}} & & \multicolumn{3}{c}{\multirow{1}{*}{Open-set lang.}} & & \multicolumn{3}{c}{\multirow{1}{*}{Open-set full}} \\
      \cmidrule{2-4}
      \cmidrule{6-8}
      \cmidrule{10-12}
      \cmidrule{14-16}
    &  \multirow{1}{*}{mAP} & \multirow{1}{*}{AUC} & \multirow{1}{*}{acc} & & \multirow{1}{*}{mAP} & \multirow{1}{*}{AUC} & \multirow{1}{*}{acc} & & \multirow{1}{*}{mAP} & \multirow{1}{*}{AUC} & \multirow{1}{*}{acc} & & \multirow{1}{*}{mAP} & \multirow{1}{*}{AUC} & \multirow{1}{*}{acc} \\
    \midrule
    
    TALL~\cite{xu-ICCV-2023} & $0.84$ & $0.85$ & $75.61$ & & $0.67$ & $0.73$ & $62.40$ & & $0.77$ & $0.78$ & $70.14$ & & $0.69$ & $0.72$ & $63.86$   \\
    MRDF~\cite{Zou-ICASSP-2024} &  $0.90$ & $0.92$  & $79.55$ & & $0.75$ & $0.81$ & 
    $67.86$ & & $0.88$ & $0.89$ & $76.91$ & & $0.80$ & $0.83$ & $70.95$  \\
    AVFF~\cite{oorloff-CVPR-2024}&  $0.97$ & $0.97$ & \textcolor{second}{$90.71$} & & $0.93$ & \textcolor{second}{$0.94$} & 
    $85.10$ & & $0.93$ & $0.93$ & $\textcolor{second}{86.58}$ & & $0.91$ & $0.93$ & \textcolor{second}{$84.85$} \\
    AVH-Align~\cite{Smeu-CVPR-2025} & $0.56$ & $0.56$ & $52.48$ & & $0.51$ & $0.51$ & $62.66$ & & $0.57$ & $0.57$ & $58.11$ & & $0.54$ & $0.54$ & $57.78$ \\
     AVH-Align$_{\text{sup}}$~\cite{Smeu-CVPR-2025} 
    & $0.81$ & $0.83$ & $75.56$ & & $0.69$ & $0.74$ & $63.84$ & & $0.76$ & $0.78$  & $72.75$ & & $0.70$ & $0.73$ & $64.90$ \\
    \midrule
    DF-Linear &  $\textcolor{second}{0.97}$&  $\textcolor{second}{0.97}$ &  $90.57$ & &  $\textcolor{second}{0.93}$ &  $0.93$ &  $\textcolor{second}{85.28}$ & &  $\textcolor{second}{0.94}$ &  $\textcolor{second}{0.94}$ & 84.06 & &  $\textcolor{second}{0.93}$ &  $\textcolor{second}{0.93}$ &  $83.98$ \\ 
    DF-MoE (ours) & $\mathbf{\textcolor{best}{0.99}}$ & $\mathbf{\textcolor{best}{0.99}}$ & $\mathbf{\textcolor{best}{97.73}}$ & & $\mathbf{\textcolor{best}{0.98}}$ & $\mathbf{\textcolor{best}{0.98}}$ & $\mathbf{\textcolor{best}{93.64}}$ & & $\mathbf{\textcolor{best}{0.99}}$ & $\mathbf{\textcolor{best}{0.99}}$ & $\mathbf{\textcolor{best}{94.55}}$ & & $\mathbf{\textcolor{best}{0.98}}$ & $\mathbf{\textcolor{best}{0.98}}$ &	$\mathbf{\textcolor{best}{92.95}}$ \\
    \bottomrule
  \end{tabular}
  }
  \vspace{0.15cm}
  \caption{Results on MAVOS-DD obtained by TALL \cite{xu-ICCV-2023}, MRDF~\cite{Zou-ICASSP-2024}, AVFF~\cite{oorloff-CVPR-2024}, AVH-Align~\cite{Smeu-CVPR-2025}, AVH-Align$_\text{sup}$~\cite{Smeu-CVPR-2025}, DF-Linear, and our DF-MoE. The best performing method is highlighted in \textbf{\textcolor{best}{blue bold}}, and the second-best in \textcolor{second}{orange}. Our DF-MoE outperforms all previous state-of-the-art methods, regardless of the evaluation setup.}
  \label{table-results}
  \vspace{-0.1cm}
\end{table}

\begin{table}[t!]
  \centering
  \setlength\tabcolsep{0.1cm}
  \fontsize{8}{9}\selectfont{
  \begin{tabular}{lccc}
    \toprule
      \multirow{2.5}{*}{Method} &  \multicolumn{3}{c}{\multirow{1}{*}{AVLips}}  \\
      \cmidrule{2-4}
    &  \multirow{1}{*}{mAP} & \multirow{1}{*}{AUC} & \multirow{1}{*}{acc}   \\
    \midrule
     RealForensics~\cite{haliassos-CVPR-2022} & $0.90$ & - & $91.78$  \\ 
     LipForensics~\cite{haliassos-CVPR-2021} & $ 0.82$ & - & $86.13$ \\ 
     LipFD~\cite{Liu-NeurIPS-2024} & \textcolor{second}{$0.93$} & \textcolor{second}{$0.95$} & $\mathbf{\textcolor{best}{95.27}}$ \\
     AVH-Align~\cite{Smeu-CVPR-2025} & - & $0.89$ & - \\
     AVFF~\cite{oorloff-CVPR-2024} & $0.89$ & $0.89$ & $76.64$  \\
    \midrule
    DF-MoE (ours) & $\mathbf{\textcolor{best}{0.97}}$ & $\mathbf{\textcolor{best}{0.97}}$ & \textcolor{second}{$93.77$} \\ 
    \bottomrule
  \end{tabular}
  }
    \vspace{0.15cm}
  \caption{In-domain results on AVLips~\cite{Liu-NeurIPS-2024} obtained by state-of-the-art models vs.~DF-MoE. The best performing method is highlighted in \textbf{\textcolor{best}{blue bold}}, and the second-best in \textcolor{second}{orange}. Our DF-MoE outperforms all previous state-of-the-art methods in terms of mAP and AUC.}
  \label{table-results-avlips}
   \vspace{-0.25cm}
\end{table}

\noindent
\textbf{Evaluation measures.}
We report mean average precision (mAP), area under the ROC curve (AUC), and accuracy (acc).

\noindent
\textbf{In-domain results.}
In Table~\ref{table-results}, we present the results of DF-MoE on all four MAVOS-DD evaluation scenarios.  We observe that DF-MoE consistently yields better performance across every setup. Remarkably, DF-MoE exhibits significantly greater robustness in the open-set model scenario compared with the strongest competitor (AVFF), maintaining higher performance stability due to our integration of multiple high-level cues. 
Overall, feature diversity is a strong point of our method, as it helps the detection model to observe different failure cases of the generative models. Our MoE-based architecture has an important role in increasing the robustness to unseen generative methods, being capable of correctly balancing complementary high-level cues to achieve substantial performance boost in the open-set model setup.
In Table \ref{table-results-avlips}, we present the results of DF-MoE on AVLips~\cite{Liu-NeurIPS-2024}. 
The results demonstrate the superior performance of DF-MoE in terms of both mAP and AUC. Notably, while AVFF is an important component of our architecture, it yields significantly lower results when evaluated in isolation. This performance gap further underscores the importance of integrating highly diverse features via DF-MoE. While LipFD achieves higher accuracy than our method on AVLips, it is specifically designed to detect temporal inconsistencies between lip movements and audio (making it specifically suitable for AVLips), whereas DF-MoE provides a more general framework for multimodal deepfake detection.

 \begin{table}[t!]
    \centering
    \setlength\tabcolsep{0.14cm}
    \fontsize{8}{9}\selectfont{
    \begin{tabular}{ccccccccc}
    \toprule
         \multirow{2}{*}{rPPG} &  Face &  \multirow{2}{*}{HP+Gaze} & AV & AV & AVFF+ & \multirow{2}{*}{mAP} & \multirow{2}{*}{AUC} & \multirow{2}{*}{acc}\\
          & segmentation &  & emotion & transformer & Effort &\\
             \midrule
          \textcolor{ForestGreen}{\checkmark} & \textcolor{Red}{\xmark} & \textcolor{Red}{\xmark} & \textcolor{Red}{\xmark} & \textcolor{Red}{\xmark} & \textcolor{Red}{\xmark} & $0.73$ & $0.74$ & $66.18$ \\
           \textcolor{Red}{\xmark} & \textcolor{ForestGreen}{\checkmark}  & \textcolor{Red}{\xmark} & \textcolor{Red}{\xmark} & \textcolor{Red}{\xmark} & \textcolor{Red}{\xmark} & $0.73$ & $0.72$ & $66.58$ \\
           \textcolor{Red}{\xmark}  & \textcolor{Red}{\xmark} & \textcolor{ForestGreen}{\checkmark} & \textcolor{Red}{\xmark} & \textcolor{Red}{\xmark} & \textcolor{Red}{\xmark} & $0.66$ & $0.69$ & $62.51$ \\
           \textcolor{Red}{\xmark}  & \textcolor{Red}{\xmark} &  \textcolor{Red}{\xmark} & \textcolor{ForestGreen}{\checkmark} &\textcolor{Red}{\xmark} & \textcolor{Red}{\xmark} & $0.78$ & $0.78$ & $70.48$\\
           \textcolor{Red}{\xmark}  & \textcolor{Red}{\xmark} &  \textcolor{Red}{\xmark}  &\textcolor{Red}{\xmark} & \textcolor{ForestGreen}{\checkmark} & \textcolor{Red}{\xmark} & $0.88$ &	$0.88$& $81.40$ \\
           \textcolor{Red}{\xmark}  & \textcolor{Red}{\xmark} &  \textcolor{Red}{\xmark}  &\textcolor{Red}{\xmark}  & \textcolor{Red}{\xmark} & \textcolor{ForestGreen}{\checkmark} & $0.96$ & $0.96$ & $88.58$ \\
            \midrule
           \textcolor{ForestGreen}{\checkmark}& \textcolor{ForestGreen}{\checkmark}& \textcolor{Red}{\xmark} &  \textcolor{Red}{\xmark}  &\textcolor{Red}{\xmark} & \textcolor{Red}{\xmark} & $0.74$ & $0.74$ &	$67.33$ \\
           \textcolor{ForestGreen}{\checkmark}& \textcolor{ForestGreen}{\checkmark}& \textcolor{ForestGreen}{\checkmark} &  \textcolor{Red}{\xmark}  &\textcolor{Red}{\xmark} & \textcolor{Red}{\xmark} & $0.74$ & $0.75$ & $68.28$ \\
           \textcolor{ForestGreen}{\checkmark}& \textcolor{ForestGreen}{\checkmark}& \textcolor{ForestGreen}{\checkmark} &  \textcolor{ForestGreen}{\checkmark}  &\textcolor{Red}{\xmark} & \textcolor{Red}{\xmark} & $0.78$ & $0.78$ & $68.18$ \\
           \textcolor{ForestGreen}{\checkmark}& \textcolor{ForestGreen}{\checkmark}& \textcolor{ForestGreen}{\checkmark} &  \textcolor{ForestGreen}{\checkmark}  &\textcolor{ForestGreen}{\checkmark} & \textcolor{Red}{\xmark} & 0.92 & 0.92 & 85.85
           \\
           \textcolor{ForestGreen}{\checkmark}& \textcolor{ForestGreen}{\checkmark}& \textcolor{ForestGreen}{\checkmark} &  \textcolor{ForestGreen}{\checkmark}  &\textcolor{ForestGreen}{\checkmark} & \textcolor{ForestGreen}{\checkmark} & $0.98$ & $0.98$ & $92.95$\\
    \bottomrule
    \end{tabular}
    }
      \vspace{0.15cm}
\caption{Ablation study on the impact of each feature type on the final performance of DF-MoE on MAVOS-DD (open-set full). The best individual components are AV Transformer and AVFF. However, the complementary cues brought by the other pre-trained models (HP+Gaze, rPPG, audio-visual emotion, face segmentation) bring consistent performance gains, all contributing to the final performance of DF-MoE.}
\label{tab:ablation}
 \vspace{-0.25cm}
\end{table}

\noindent
\textbf{Ablation studies.} To validate the utility of the sparse MoE, we compare DF-MoE with DF-Linear in Tables \ref{table-results} and Table~\ref{tab:cross}. DF-Linear replaces the MoE module with a linear layer applied on the concatenation of all high-level features. While DF-Linear generally surpasses state-of-the-art detectors in the in-domain setting, its in-domain performance is consistently below our DF-MoE. Significant gaps are also observed for the cross-domain evaluation on PolyGlotFake and Vox+FakeAVCeleb.

In Table~\ref{tab:ablation}, we present a comprehensive ablation study designed to isolate and quantitatively evaluate the contribution of each individual feature type to the final performance. This analysis is conducted on the open-set full setup of MAVOS-DD. The ablation results indicate that the most important representations are extracted by the two audio-visual models (AV Transformer and AVFF), which already surpass some of the state-of-the-art models. Nevertheless, every high-level cue achieves non-trivial performance, when evaluated in isolation. This individual effectiveness indicates that the proposed features capture useful information that can be further harnessed in the full pipeline. 
Gradually integrating the individual components improves performance, indicating that the high-level cues exhibit a strong complementary effect, boosting the overall robustness and performance of DF-MoE. We report more ablations in the supplementary.


\begin{table}[t!]
    \centering
    \setlength\tabcolsep{0.1cm}
    \fontsize{8}{9}\selectfont{
    \begin{tabular}{clccc c ccc c ccc}
    \toprule
        \multirow{2.5}{*}{Year} & \multirow{2.5}{*}{Method} 
         &\multicolumn{3}{c}{PolyGlotFake~\cite{Hou-ICPR-2024}} & & \multicolumn{3}{c}{BioDeepAV~\cite{Croitoru-Arxiv-2024}}  & & \multicolumn{3}{c} {Vox+FakeAVCeleb~\cite{khalid-NeurIPS-2021}}\\
         \cmidrule{3-5}
         \cmidrule{7-9}
         \cmidrule{11-13}
        & & mAP & AUC & acc  & & mAP & AUC & acc & & $\,\,$mAP$\,\,$ & $\,$AUC$\,$ & acc \\
    \midrule
  \multirow{3}{*}{2018} & MesoNet~\cite{Afchar-WIFS-2018} 
    & - & $0.57$ & - & & - & - & - & & - & - & -\\
    & MesoInception~\cite{Afchar-WIFS-2018} 
     & - & $0.58$ & - & & - & - & - & & - & - & -\\
     & DSP-FWA~\cite{Li-CVPRW-2018} 
    & - & $0.67$ & - & & - & - & - & & - & - & -\\
    \midrule
    \multirow{2}{*}{2019} & XceptionNet~\cite{Roessler-ICCV-2019} 
     & - & $0.61$ & - & & - & $0.57$ & - & & - & - & -\\
    & EfficienNet-B4~\cite{tan-ICML-2019} 
     & - & $0.58$ & - & & - & - & - & & - & - & -\\
     \midrule
    \multirow{2}{*}{2020} & FFD~\cite{Dang-CVPR-2020} 
     & - & $0.60$ & - & & - & - & - & & - & - & -\\
     & F3Net~\cite{Qian-ECCV-2020} 
    & - & $0.64$ & - & &  - & $0.50$ & - & & - & - & -\\
    \midrule
   \multirow{2}{*}{2022} & CORE~\cite{Ni-CVPRW-2022} 
     & - & $0.62$ & - & & - & - & - & & - & - & -\\
   & RECCE~\cite{Cao-CVPR-2022} 
    & - & $0.66$ & - & &   - & $0.50$ & - & & - & - & -\\

        \midrule
        
    \multirow{2}{*}{2023} & UCF~\cite{Yan-ICCV-2023}& - & - & - & &  - &  $0.49$ & - & & - & - & -\\

        & TALL$^*$~\cite{xu-ICCV-2023} 
    & $0.56$ & $0.58$ & $32.74$  & & $0.83$ & $0.82$ & $74.66$& & $0.50$ & $0.57$ & $55.06$ \\

    \midrule
    
    \multirow{3}{*}{2024} & XRes~\cite{Hou-ICPR-2024}
    & - & $0.68$ & - & & - & - & - & & - & - & -\\

    & AVFF$^*$~\cite{oorloff-CVPR-2024} 
    & $\textcolor{second}{0.88}$ & $0.91$ & $\textcolor{second}{97.93}$ & & $\textcolor{second}{0.95}$ &$\textcolor{second}{0.95}$ & $70.78$ & & ${\textcolor{second}{0.69}}$ & ${\textcolor{second}{0.70}}$ & $55.56$ \\
    
    & MRDF$^*$~\cite{Zou-ICASSP-2024} 
    & $0.51$ & $0.39$ & $5.56$ & & $0.53$ & $0.53$ & $52.12$ & & $0.65$ & $0.65$ & ${\textcolor{second}{61.83}}$\\

    \midrule
    \multirow{3}{*}{2025} & StA~\cite{Yan-CVPR-2025} & - & - & - & & - & $0.62$ & - & & - & - & -\\
    & ForAda~\cite{Cui-CVPR-2025} & - & $0.87$ & - & & - & - & - & & - & - & -\\
    & Effort~\cite{Yan-ICML-2025} & - & $0.85$ & - & & - & - & - & & - & - & -\\

    \midrule
    \multirow{5.5}{*}{2026} & GenD (CLIP)~\cite{Yermakov-WACV-2026} & - & $0.90$ & - & & - & - & - & & - & - & -\\
    & GenD (PE)~\cite{Yermakov-WACV-2026} & - & $0.92$ & - & & - & - & - & & - & - & -\\
    & GenD (DINO)~\cite{Yermakov-WACV-2026} & - & $\textcolor{second}{0.92}$ & - & & - & - & - & & - & - & - \\
        
    \cmidrule{2-13}
    & DF-Linear & $0.82$ & $0.89$ & $94.57$ & & $\mathbf{\textcolor{best}{0.99}}$ & $\mathbf{\textcolor{best}{0.99}}$ & $\mathbf{\textcolor{best}{97.87}}$ & & $0.46$	& $0.45$ & $50.91$ \\
     & DF-MoE (ours) 
    & $\mathbf{\textcolor{best}{0.93}}$ & $\mathbf{\textcolor{best}{0.94}}$ & $\mathbf{\textcolor{best}{98.49}}$ & & $\mathbf{\textcolor{best}{0.99}}$ & $\mathbf{\textcolor{best}{0.99}}$ & ${\textcolor{second}{96.77}}$ & & $\mathbf{\textcolor{best}{0.86}}$	& $\mathbf{\textcolor{best}{0.88}}$ & $\mathbf{\textcolor{best}{81.59}}$ \\
    \bottomrule
    \end{tabular}
}
  \vspace{0.15cm}
\caption{Cross-dataset evaluation on PolyGlotFake, BioDeepAV and Vox+FakeAVCeleb. The best performing method is highlighted in \textbf{\textcolor{best}{blue bold}}, and the second-best in \textcolor{second}{orange}. DF-MoE obtains the best performance on both datasets in terms of AUC and acc. Results of methods marked with an asterisk are reproduced using publicly available code.}
\label{tab:cross}
 \vspace{-0.25cm}
\end{table}

\noindent
\textbf{Cross-dataset results.} Next, we verify the robustness of DF-MoE to different manipulation methods and various real-world data sources. Specifically, we report results on PolyGlotFake, BioDeepAV and Vox+FakeAVCeleb in Table~\ref{tab:cross}, using MAVOS-DD and AVLips as training data. The results follow the same trend observed in the open-set scenarios of MAVOS-DD, namely that  DF-MoE obtains state-of-the-art results on all three datasets in terms of mAP and AUC. We compare these results with several other methods. We compute the performance metrics of AVFF, TALL and MRDF using the code available in the corresponding public repositories. For the remaining methods, the AUC metric is taken from the official publications. While some of the evaluated methods~\cite{Yan-CVPR-2025, Yan-ICCV-2023} are designed to improve generalization in deepfake detection, they still struggle to maintain optimal performance when encountering the significant data distribution shifts in PolyGlotFake, BioDeepAV and Vox+FakeAVCeleb.

\begin{figure}[!t]
    \centering
    \includegraphics[width=0.95\linewidth]{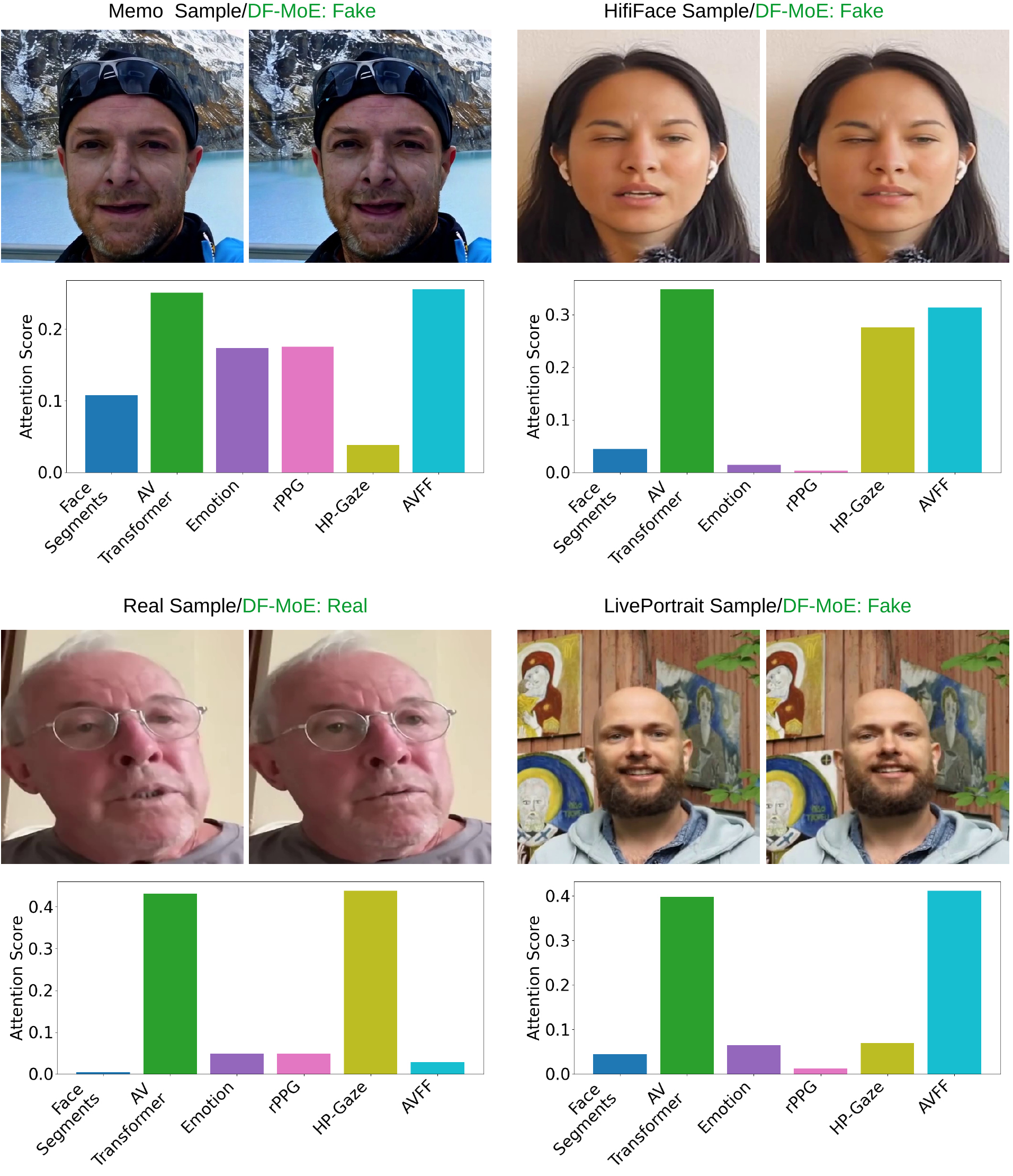}
    \vspace{-0.25cm}
    \caption{Examples of videos correctly classified by DF-MoE, alongside the corresponding attention scores of each high-level cue. Best viewed in color.} 
    \label{fig:qualitative_results_good}
    \vspace{-0.4cm}
\end{figure}

\noindent
\textbf{Qualitative analysis.} In Figure~\ref{fig:qualitative_results_good}, we show four videos from the MAVOS-DD test set that are correctly classified by DF-MoE. The fake videos are generated by three of the included manipulation methods (Memo, HifiFace and LivePortrait). These three methods cover all the visual manipulation types available in MAVOS-DD, namely talking face synthesis (Memo), face swapping (HifiFace) and lip synchronization (LivePortrait). Along with the video frames, we illustrate the attention scores associated with each high-level cue. The scores are computed based on the attention weights provided by the multi-head attention layer. To obtain a single value for each feature, given the attention weights, we compute their average across the head and query dimensions.

By analyzing the results shown in Figure~\ref{fig:qualitative_results_good}, we can make several interesting observations. The AV Transformer and AVFF models have generally high contributions, essentially due to their strong individual performance coming from the direct analysis of manipulated audio-visual content. The top-left example shows a slight skin tone variation between frames, which explains the high rPPG score. In the top-right example, head pose and gaze receive high importance, likely capturing gaze inconsistencies caused by face-swapping artifacts around the eyes. In the bottom-right LivePortrait sample, where lip movements are altered, the AV Transformer and AVFF effectively detect the manipulation by assessing audio-visual synchronization. Lastly, in the real sample (bottom-left), high scores from the AV Transformer and head pose features indicate that the model correctly identifies natural audio-visual synchronization and authentic head movements.

\noindent
\textbf{Computational complexity.}
Our complete pipeline, which integrates all pre-trained models, has 508.18 GFLOPs and 424.14M parameters. During training, we only update 19.9M (5\%) of the 424.14M inference parameters. In Table~\ref{tab:inference_time}, we report the average inference time in seconds for all stages of our pipeline. 
With all the components in place, DF-MoE reaches near real-time processing speed at inference, on both one or two GPUs. Training is conducted using two RTX 5090 GPUs. DF-MoE is trained for 10 epochs at roughly 5 hours and 20 minutes per epoch, totaling approximately 52 hours of compute time per experiment.



\begin{table}[t!]
    \centering
        \setlength\tabcolsep{0.2cm}
    \fontsize{8}{9}\selectfont{
    \begin{tabular}{cccccc}
    \toprule
       \multirow{2}{*}{\#GPUs} & Average   & Face detection \& & Feature 
         & \multirow{2}{*}{MoE} & \multirow{2}{*}{Total} \\
         
       &  video duration  & tracking  & extraction & & \\
             \midrule
       1 & 10.97  & 3.35 & 8.47 & 1.16 & 12.98\\
       2 & 10.97  & 3.35 & 4.65 & 1.16 & 9.16\\
    \bottomrule
    \end{tabular}
}
\vspace{0.15cm}
\caption{Average video duration and stage-wise inference times (in seconds) for DF-MoE. The values are estimated across 20 videos on a machine with $1\times$AMD Ryzen Threadripper 9960X 24-core CPU and $2\times$Nvidia RTX 5090 GPU (32GB VRAM).}
\label{tab:inference_time}
\vspace{-0.25cm}
\end{table}

\vspace{-0.2cm}
\section{Conclusion}
\vspace{-0.1cm}

In this work, we addressed the out-of-domain generalization issue of audio-video deepfake detectors by proposing DF-MoE. Our framework integrates audio-visual cues provided by several pre-trained models, including head pose, gaze, face segmentation maps, rPPG signals, face and audio emotions, etc. Using pre-trained models prevents overfitting to a specific deepfake generator or dataset. Our DF-MoE also includes trainable mixture-of-experts, improving deepfake detection with their ability to create specialized features (in our case, for different input signals), while preserving the generalization capacity of our framework. The ablation results demonstrated that each high-level cue provides useful information for deepfake detection. DF-MoE obtained state-of-the-art results on three challenging datasets, outperforming the previous methods in terms of relevant performance metrics. We also demonstrated that introducing frozen pre-trained models into an efficient architecture provides state-of-the-art cross-domain performance. In future work, employing additional pre-trained models to extract complementary signals could further boost performance on deepfake detection, and even on other complex tasks, such as video planning.

\vspace{-0.2cm}
\section{Acknowledgments}
\vspace{-0.1cm}

This work was supported by a grant of the Ministry of Research, Innovation and Digitization, CCCDI - UEFISCDI, project number PN-IV-P6-6.3-SOL-2024-2-0227, within PNCDI IV. This work was in part supported by the BMFTR (FKZ: 16IS24060), and the DFG (SFB 1233, project number: 276693517).

\bibliography{main}

\setcounter{table}{0}
\renewcommand{\thetable}{A\arabic{table}}

\setcounter{figure}{0}

\renewcommand{\topfraction}{1.0}
\renewcommand{\textfraction}{0.0}

\renewcommand{\floatpagefraction}{1.0}

\section{Supplementary}

\subsection{Additional Ablation Studies}

\noindent
\textbf{Ablation of hyperparameters.} In Table~\ref{tab:ablation}, we present an ablation study evaluating two key hyperparameters of DF-MoE, the expert dropout ratio ($d$) and the number of experts ($n$). To prevent overfitting on the MAVOS-DD dataset, we limit our exploration of these parameters to a small set of values. The results confirm that the values used in our main experiments ($d=0.2$ and $n=6$) yield optimal performance on the open-set full scenario of MAVOS-DD. Moreover, all explored versions significantly surpass the state-of-the-art competitors \cite{oorloff-CVPR-2024,Smeu-CVPR-2025,xu-ICCV-2023,Zou-ICASSP-2024} (see Table 1 from the main paper), indicating that DF-MoE attains consistently high performance, even with suboptimal hyperparameter configurations.

\begin{table}[t]
    \centering
    \setlength\tabcolsep{0.14cm}
    \fontsize{8}{9}\selectfont{
    \begin{tabular}{ccccc}
    \toprule
         Dropout Ratio &  \#Experts &  mAP & AUC & acc\\
             \midrule
          $0.2$ & $6$ & $0.98$ & $0.98$ & $92.95$\\
        $0.4$ & $6 $& $0.97 $&	$0.97 $& $91.70$ \\
        $0.2$ & $12$ &$ 0.96$ &$0.97$ &$ 92.54$ \\
         $0.4$ & $12$ &$ 0.96$ &$0.96$ & $91.24$ \\
    \bottomrule\\
    \end{tabular}
    }
    \caption{Ablation study on the expert dropout ratio ($d$) and the number of experts ($n$) included in DF-MoE. The study is conducted on the open-set full protocol of MAVOS-DD \cite{Croitoru-ArXiv-2025}.}
    \label{tab:ablation}
\end{table}

\begin{figure}[t]
\centering
\begin{subfigure}[t!]{.9\textwidth}
  \centering
\includegraphics[width=.99\textwidth]{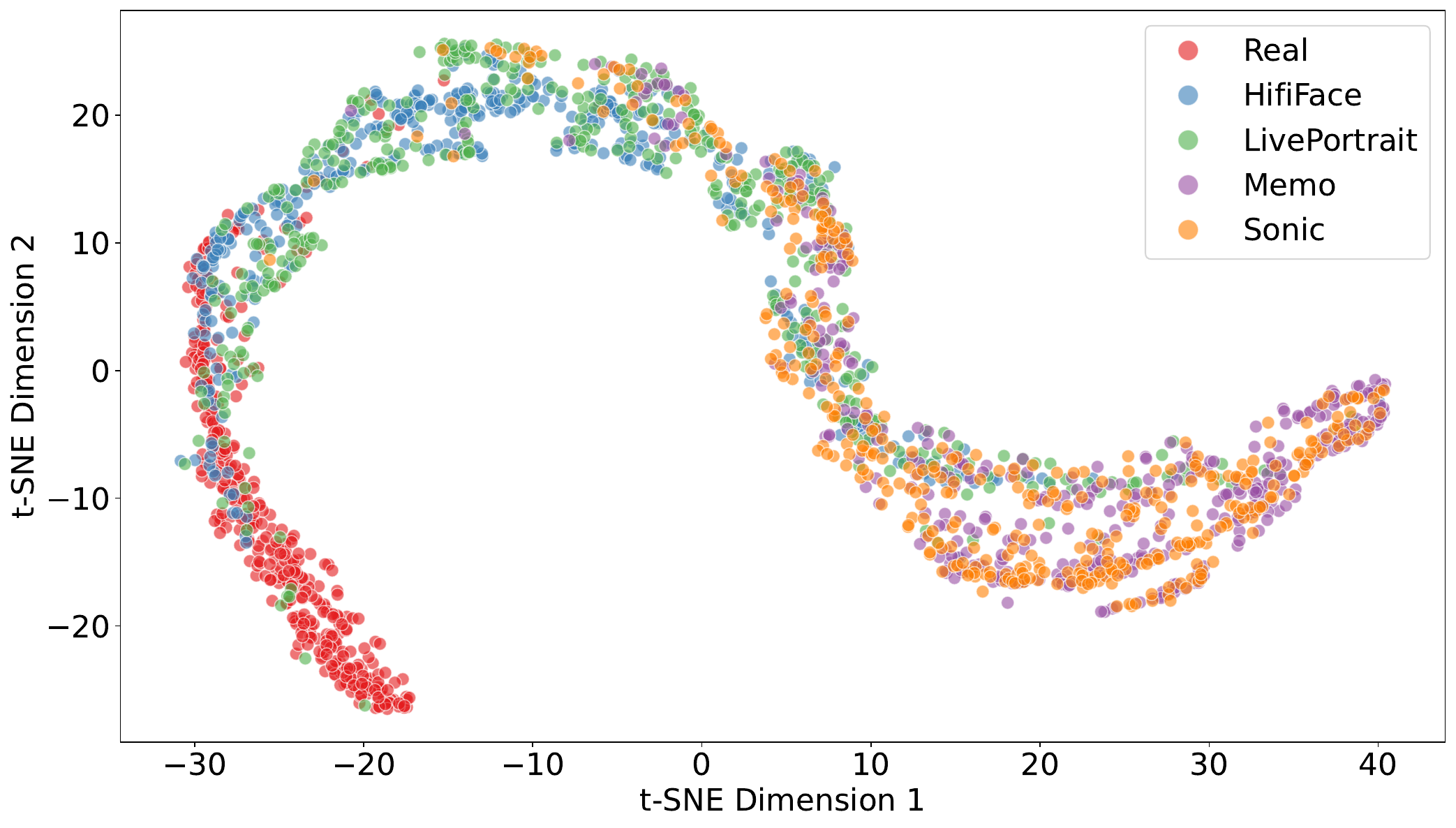} 
\vspace{-0.2cm}
\caption{Latent space of DF-MoE before introducing the CRO loss.}
\label{fig:without_cro}
\vspace{0.2cm}
\end{subfigure}
\begin{subfigure}[t!]{.9\textwidth}
  \centering
  \includegraphics[width=.99\linewidth]{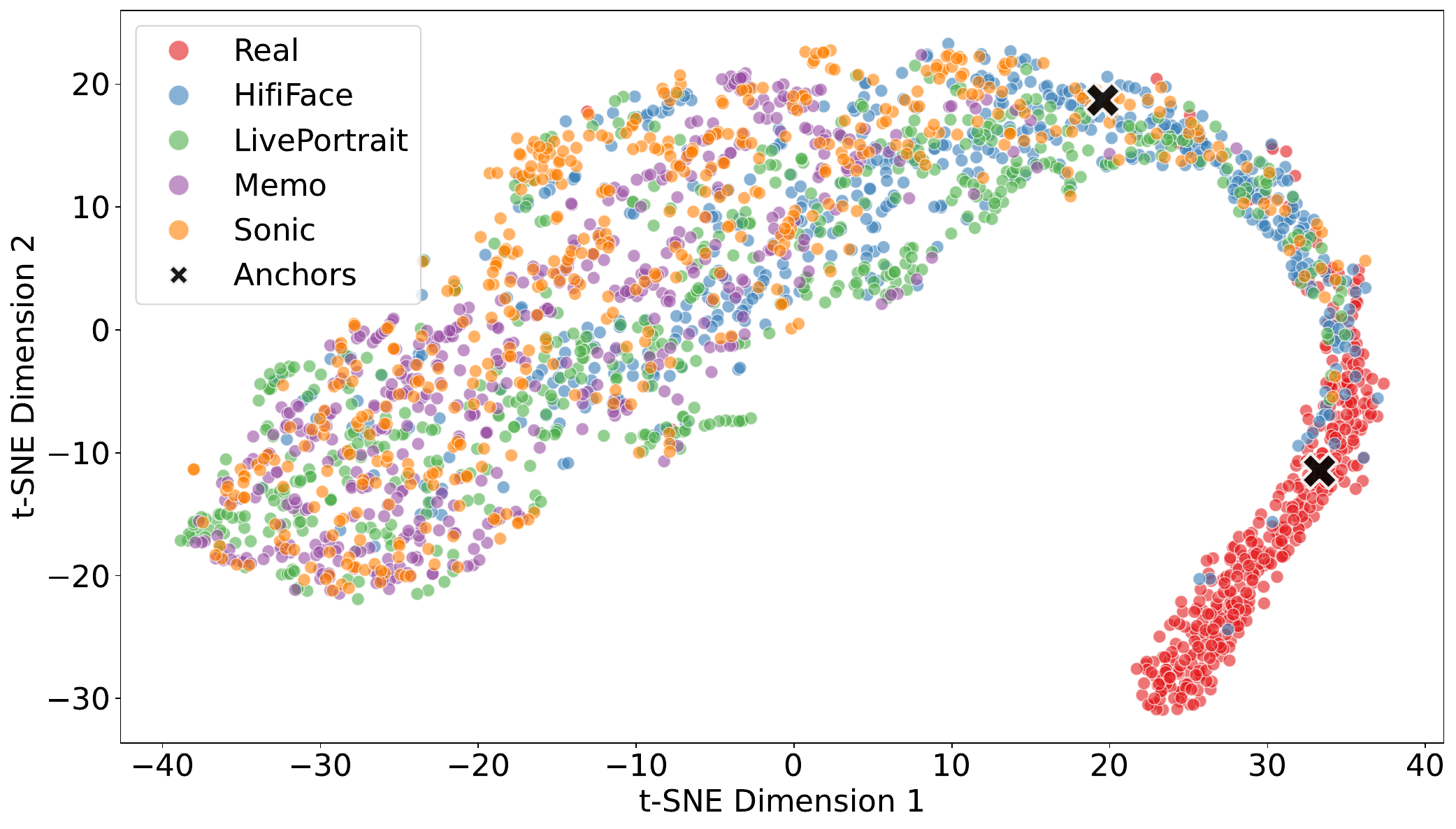}  
  \vspace{-0.2cm}
  \caption{Latent space of DF-MoE after introducing the CRO loss.}
  \label{fig:with_cro}
  \vspace{0.3cm}
\end{subfigure}
\caption{Comparison between latent spaces of DF-MoE, before and after introducing CRO loss. In both cases, the model is trained on MAVOS-DD. Real samples and deepfakes generated by various methods are illustrated through different colors. Sonic and HifiFace are generative methods that do not belong to the training set. Best viewed in color.}
\label{fig:tsne}
\end{figure}

\noindent
\textbf{Effect of contractive-repulsive objective.}
To showcase the effect of the CRO loss on the latent space, we present t-SNE visualizations of real and deepfake embeddings from the latent space of DF-MoE, before and after introducing the proposed CRO loss. In Figure \ref{fig:tsne}, we compare the latent space obtained after training on MAVOS-DD. Upon introducing our contractive-repulsive objective, we observe that deepfakes generated by different methods are entangled in a more compact region of the latent space. Interestingly, out-of-domain (open-set) deepfake generation methods, such as Sonic \cite{Ji-CVPR-2025} and HifiFace \cite{Wang-IJCAI-2021}, share the same behavior as in-domain (closed-set) deepfake generators.

In contrast, DF-MoE without our CRO loss spreads samples from different detection methods in a wider area, and different generative methods are located in distinctive regions, harming generalization capacity. Consequently, several deepfake samples produced by HifiFace \cite{Wang-IJCAI-2021} are entangled with the real samples. In general, real and deepfake entanglements in the latent space inherently lead to performance  degradation in the cross-dataset scenario (see Table \ref{tab:cro}). 

In summary, the t-SNE visualizations depicted in Figure \ref{fig:tsne} indicate that the latent space organization induced by our CRO loss contributes to the generalization of DF-MoE.

\begin{table}[t]
    \centering
    \setlength\tabcolsep{0.14cm}
    \fontsize{8}{9}\selectfont{
    \begin{tabular}{cccc}
    \toprule
         CRO Loss &  mAP & AUC & acc\\
             \midrule
                     \textcolor{Red}{\xmark}  & $0.919$ &	$0.936$	& $98.42$ \\

                     \textcolor{ForestGreen}{\checkmark} &  $0.929$ &	$0.941$	& $98.49$\\
    \bottomrule
    \end{tabular}
    }
    \vspace{0.2cm}
    \caption{Results of DF-MoE on PolyGlotFake \cite{Hou-ICPR-2024}, before and after introducing the proposed contractive-repulsive objective (CRO).}
    \label{tab:cro}
\end{table}

\noindent
\textbf{Semantic cues contributions.} The advantages of incorporating multiple modalities beyond AVFF are most evident in the cross-dataset results from the main paper (Table~4, last column). Additionally, in Table~\ref{tab:avff features}, we break down the contribution of each modality on PolyGlotFake \cite{Hou-ICPR-2024}. Most features improve upon the AVFF+Effort baseline. Although combining audio-video emotion features with AVFF features results in a slight performance drop (fourth row), further adding HP+Gaze (last row) outperforms the combination of AVFF and HP+Gaze features (third row). This result suggests that gaze provides complementary contextual information for facial expressions through self-attention, showing the benefits of jointly modeling complementary modalities for a more robust representation for deepfake detection.

 \begin{table}[t!]
    \centering
    \setlength\tabcolsep{0.14cm}
    \fontsize{8}{9}\selectfont{
    \begin{tabular}{ccccccccc}
    \toprule
         {AVFF+} &  Face &  \multirow{2}{*}{HP+Gaze} & AV & AV & \multirow{2}{*}{rPPG} & \multirow{2}{*}{mAP} & \multirow{2}{*}{AUC} & \multirow{2}{*}{acc} \\
          Effort & segmentation &  & emotion & transformer & & & & \\
                 \midrule
          \textcolor{ForestGreen}{\checkmark} & \textcolor{Red}{\xmark} & \textcolor{Red}{\xmark} & \textcolor{Red}{\xmark} & \textcolor{Red}{\xmark} & \textcolor{Red}{\xmark} & $0.88 $ & $0.91$ & $97.93$\\
          \textcolor{ForestGreen}{\checkmark} &\textcolor{ForestGreen}{\checkmark} & \textcolor{Red}{\xmark} & \textcolor{Red}{\xmark} & \textcolor{Red}{\xmark} & \textcolor{Red}{\xmark} & $0.90$ & $0.90$ & $96.70$\\
          \textcolor{ForestGreen}{\checkmark} & \textcolor{Red}{\xmark} & \textcolor{ForestGreen}{\checkmark} & \textcolor{Red}{\xmark} & \textcolor{Red}{\xmark} & \textcolor{Red}{\xmark} & $0.91 $ & $0.92$ & $96.65$\\
          
          \textcolor{ForestGreen}{\checkmark} & \textcolor{Red}{\xmark} & \textcolor{Red}{\xmark} & \textcolor{ForestGreen}{\checkmark} & \textcolor{Red}{\xmark} & \textcolor{Red}{\xmark} & $0.87 $ & $0.88$ & $91.71$\\
          \textcolor{ForestGreen}{\checkmark} & \textcolor{Red}{\xmark} & \textcolor{Red}{\xmark} &  \textcolor{Red}{\xmark}  & \textcolor{ForestGreen}{\checkmark} & \textcolor{Red}{\xmark} & $0.92 $ & $0.93$ & $98.70$\\
          \textcolor{ForestGreen}{\checkmark} & \textcolor{Red}{\xmark} & \textcolor{Red}{\xmark} &  \textcolor{Red}{\xmark}  & \textcolor{Red}{\xmark} & \textcolor{ForestGreen}{\checkmark} &  $0.90 $ & $0.92$ & $98.56$\\
          \textcolor{ForestGreen}{\checkmark} & \textcolor{Red}{\xmark} & \textcolor{ForestGreen}{\checkmark} & \textcolor{ForestGreen}{\checkmark} & \textcolor{Red}{\xmark} & \textcolor{Red}{\xmark} & $0.93 $ & $0.93$ & $98.70$\\
    \bottomrule
    \end{tabular}
    }
      \vspace{0.2cm}
\caption{Ablation study on combinations of AVFF+Effort features and other high-level features included in DF-MoE. Results are reported on PolyGlotFake \cite{Hou-ICPR-2024}.}
\label{tab:avff features}
\end{table}




\begin{table}[t!]
    \centering
    \setlength\tabcolsep{0.14cm}
    \fontsize{8}{9}\selectfont{
    \begin{tabular}{l ccc}
    \toprule
       Method & mAP & AUC & acc \\
        \midrule
        DF-MoE (AVFF full)  &  $0.94$ & $0.94$ & $86.38$ \\
        DF-MoE (AVFF+Effort) & $0.98$ & $0.98$ & $92.95$ \\
    \bottomrule
    \end{tabular}
}
  \vspace{0.2cm}
\caption{Full vs.~parameter-efficient training of AVFF encoder on MAVOS-DD (open-set full). Fine-tuning based on Effort~\cite{Yan-ICML-2025} surpasses full fine-tuning.}
\label{tab:avff strategies}
\end{table}

\noindent
\textbf{Fine-tuning feature extractors.} Fully fine-tuning the foundational encoders is computationally impractical due to the massive memory footprint and processing costs required. Beyond these computational constraints, full fine-tuning can compromise generalization, as highly parameterized models are particularly prone to overfitting to dataset-specific forgery artifacts, and thus limiting their ability to detect unseen manipulation techniques. This phenomenon is corroborated by Table~\ref{tab:avff strategies}, which demonstrates that applying full fine-tuning on the AVFF encoder within the DF-MoE architecture actually degrades performance. In contrast, adapting the encoder using the Effort~\cite{Yan-ICML-2025} strategy yields superior results, proving that parameter-efficient fine-tuning approach balances feature adaptation with robust generalization.

\noindent
\textbf{Robustness to missing modalities.} We deliberately omit positional embeddings from our token representations, enabling DF-MoE to work seamlessly when one or more modalities are missing. To demonstrate this, we evaluate DF-MoE on the Celeb-DF (v2) \cite{Li-CVPR-2020b} dataset in the cross-dataset scenario in Table~\ref{tab:celebdf results}, where audio is not available. In this setting, DF-MoE safely ignores audio-specific features, while still obtaining competitive performance. This implies that DF-MoE is robust to changes in the concatenation order of features, and consequently, to variations in the set of available modalities.

\begin{table}[t!]
    \centering
    \setlength\tabcolsep{0.06cm}
    \fontsize{8}{9}\selectfont{
    \begin{tabular}{l c}
    \toprule
       Method & AUC \\
    \midrule
       F3Net~\cite{Qian-ECCV-2020}& 0.789\\
       CORE~\cite{Ni-CVPRW-2022} & 0.809 \\
       RECCE~\cite{Cao-CVPR-2022} & 0.823 \\
       UCF~\cite{Yan-ICCV-2023} & 0.837 \\
       LSDA~\cite{Yan-CVPR-2024} & 0.875\\
       DF-MoE (ours) & 0.834\\
    \bottomrule
    \end{tabular}
}
  \vspace{0.2cm}
\caption{Cross-dataset evaluation on Celeb-DF (v2) \cite{Li-CVPR-2020b}, a video-only dataset. The results demonstrate that DF-MoE obtains competitive results even when the audio modality is missing.}
\label{tab:celebdf results}
 \vspace{-0.55cm}
\end{table}

\begin{figure}[t]
    \centering
    \includegraphics[width=1.\linewidth]{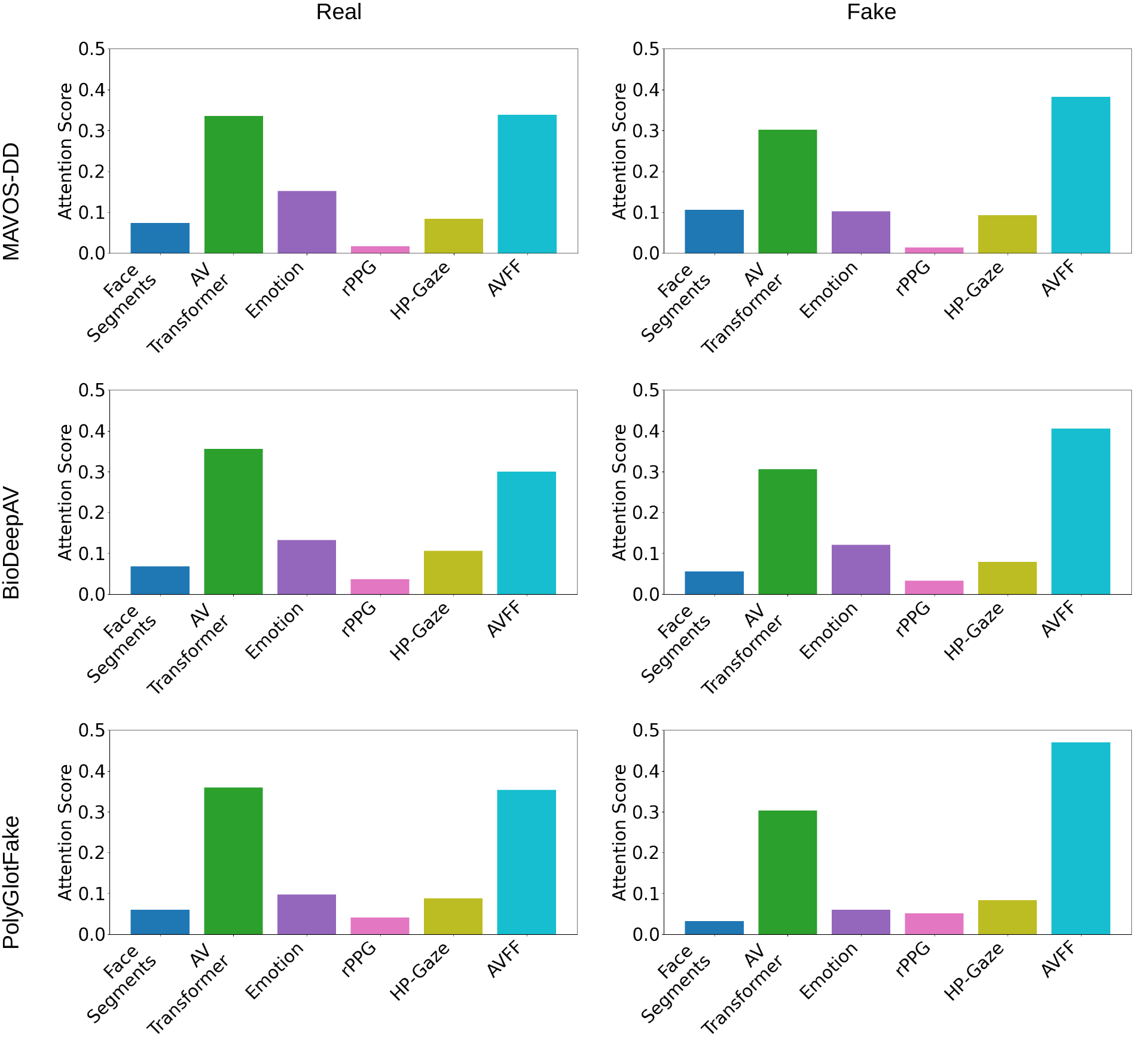}
    \vspace{-0.6cm}
    \caption{Average attention scores across datasets and class labels.}
    \label{fig:attention_scores}
\end{figure}

\begin{figure}[t]
    \centering
    \includegraphics[width=0.98\linewidth]{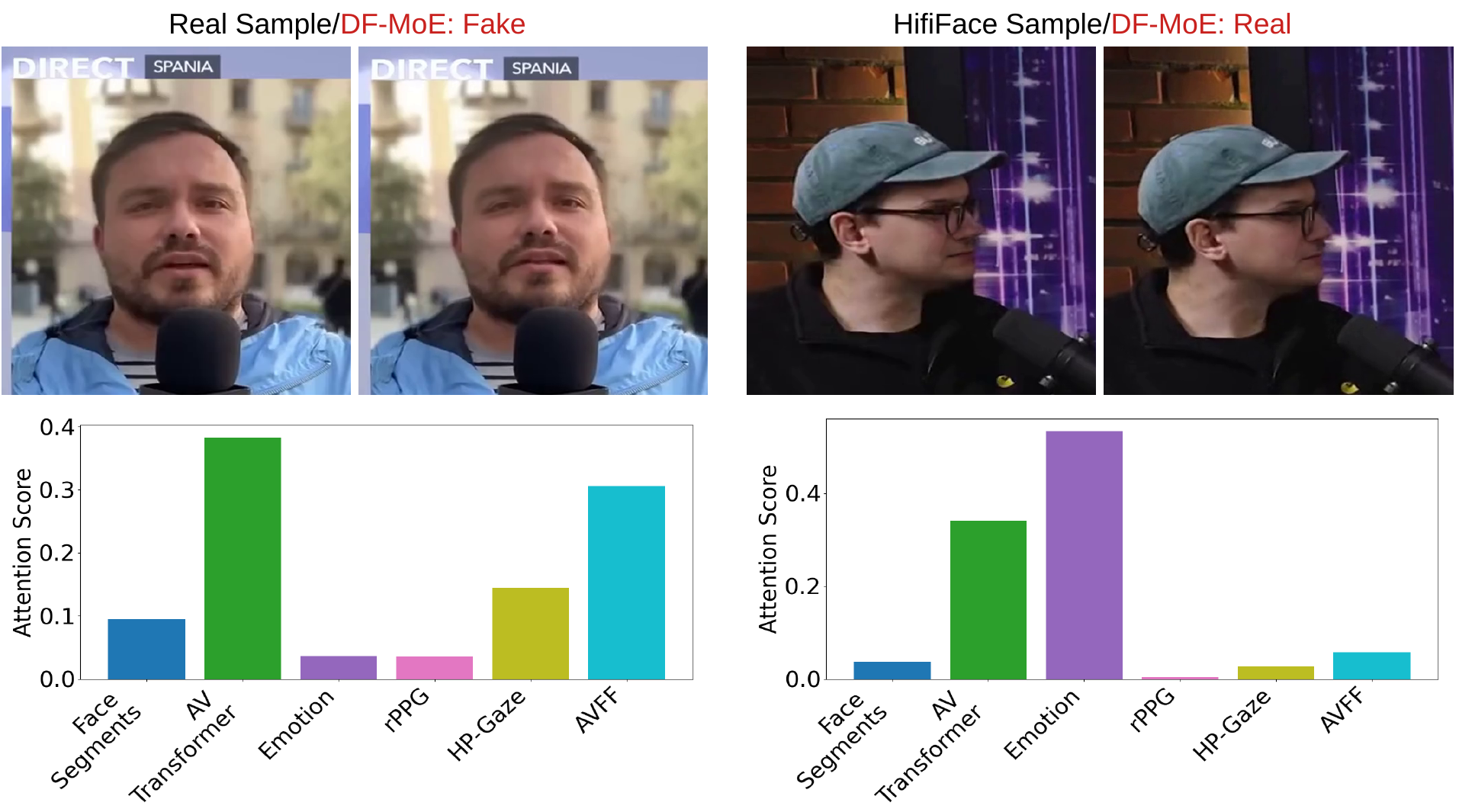}
    \vspace{-0.15cm}
    \caption{Examples of videos incorrectly classified by DF-MoE, alongside the corresponding attention scores of each feature type. Best viewed in color.} 
    \label{fig:qualitative_results_bad}
\end{figure}

\subsection{Qualitative Result Analysis}

In Figure~\ref{fig:attention_scores}, we show the average attention scores for each dataset, computed over $100$ real and $100$ fake videos randomly sampled from the respective test subsets. The results indicate that feature importance remains highly consistent across different datasets. This suggests that our model relies on universal, domain-agnostic features rather than overfitting to dataset-specific artifacts, which directly explains its strong generalization capabilities on unseen data.
\begin{figure*}[!t]
\begin{subfigure}[t!]{.40\textwidth}
  \centering
  \includegraphics[width=.8\linewidth]{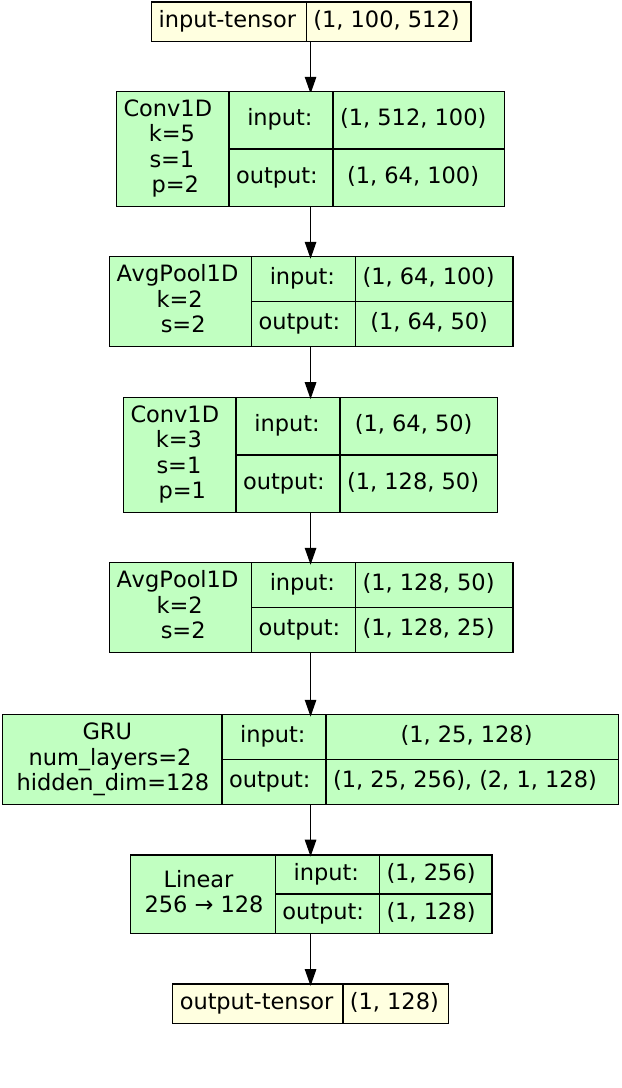}  
  \caption{rPPG.}
  \label{fig:rppg}
\end{subfigure}
\hfill
\begin{subfigure}[t!]{.56\textwidth}
  \centering
\includegraphics[width=.99\textwidth]{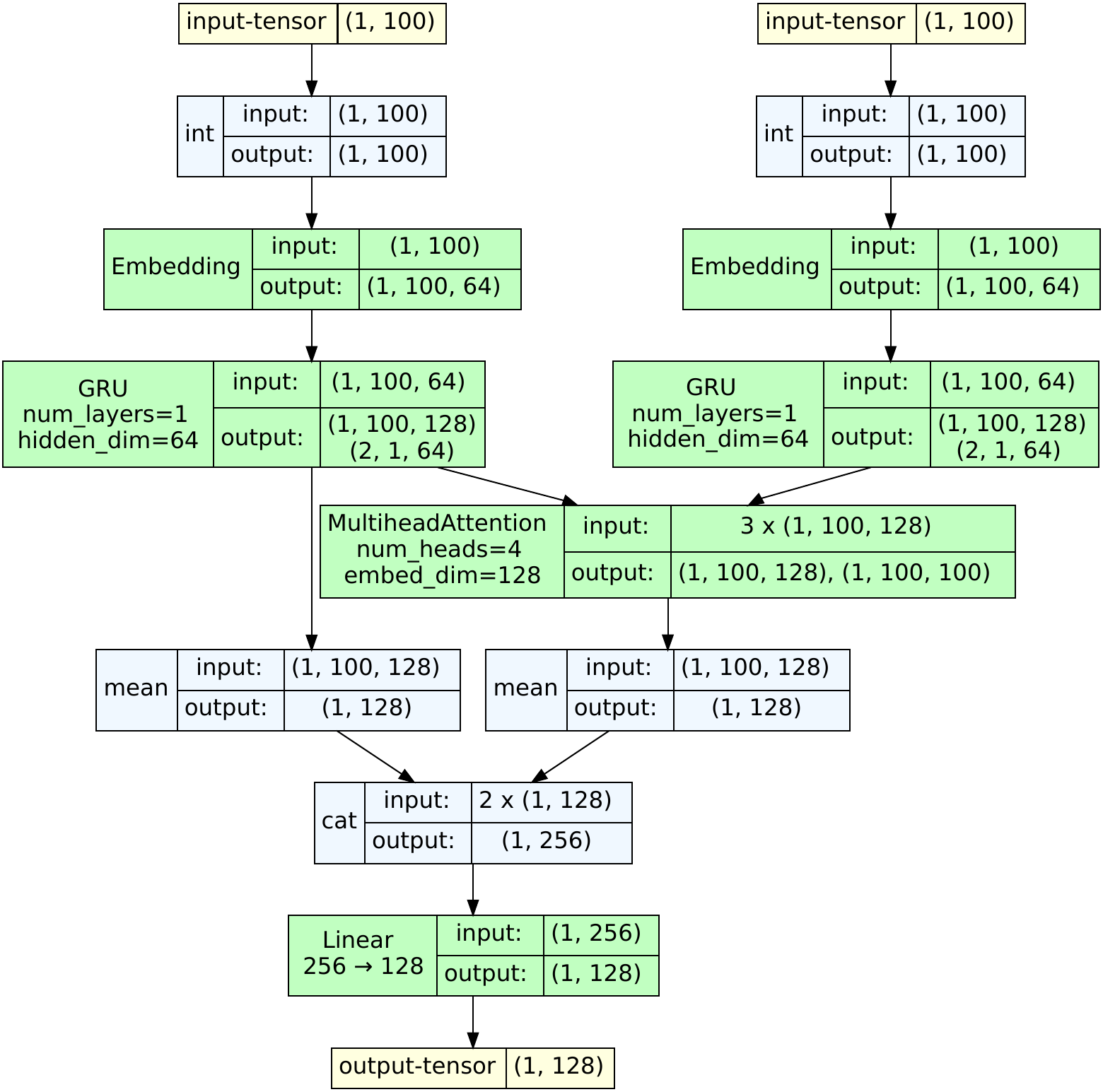} 
\caption{AV emotion.}
\label{fig:semantic}
\end{subfigure}
\vspace{0.4cm}

\begin{subfigure}[t!]{.56\textwidth}
  \centering
\includegraphics[width=.99\textwidth]{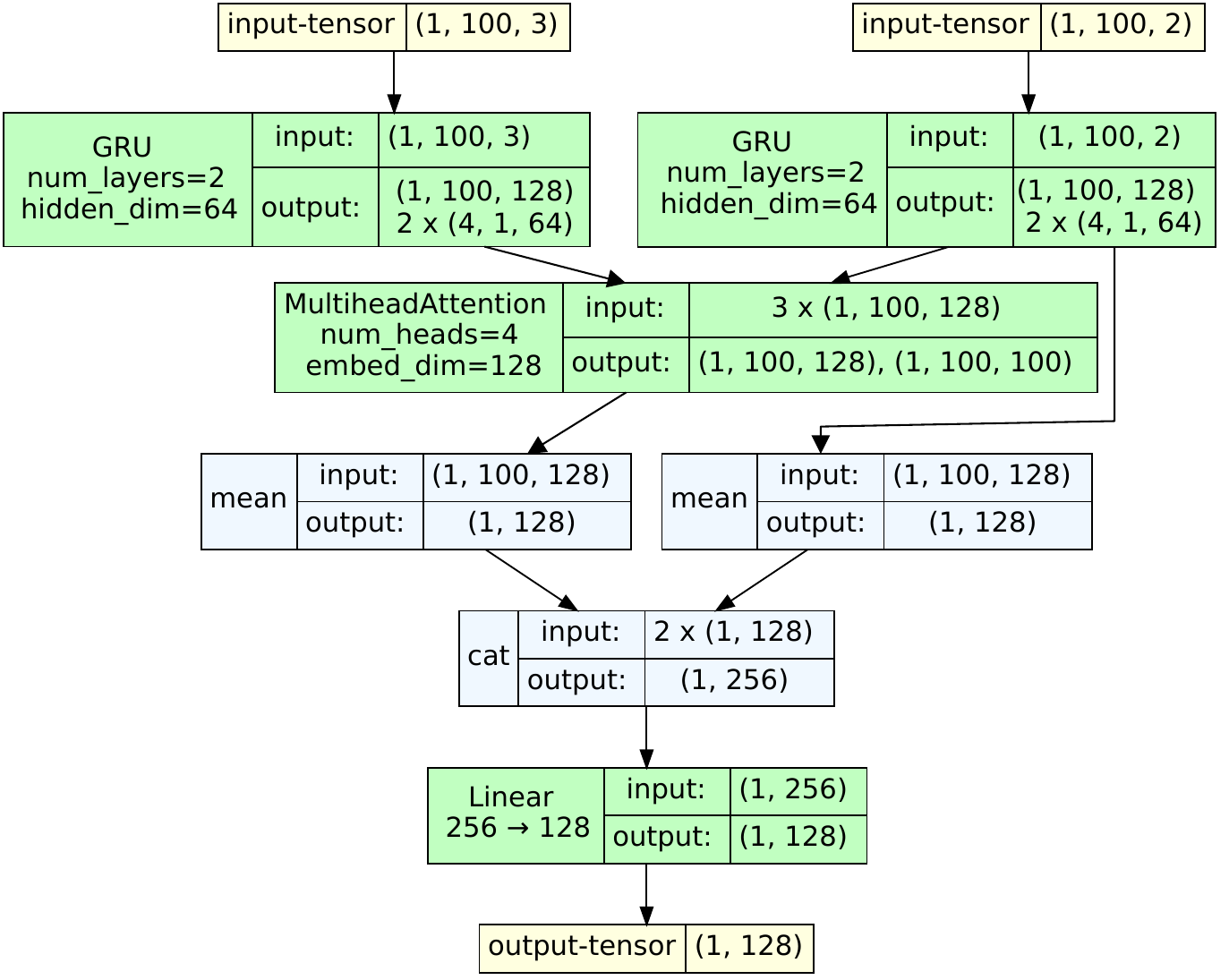} 
  \caption{HP+Gaze.}
  \label{fig:hp-gaze}
\end{subfigure}
\hfill
\begin{subfigure}[t!]{.40\textwidth}
  \centering
\includegraphics[width=.8\textwidth]{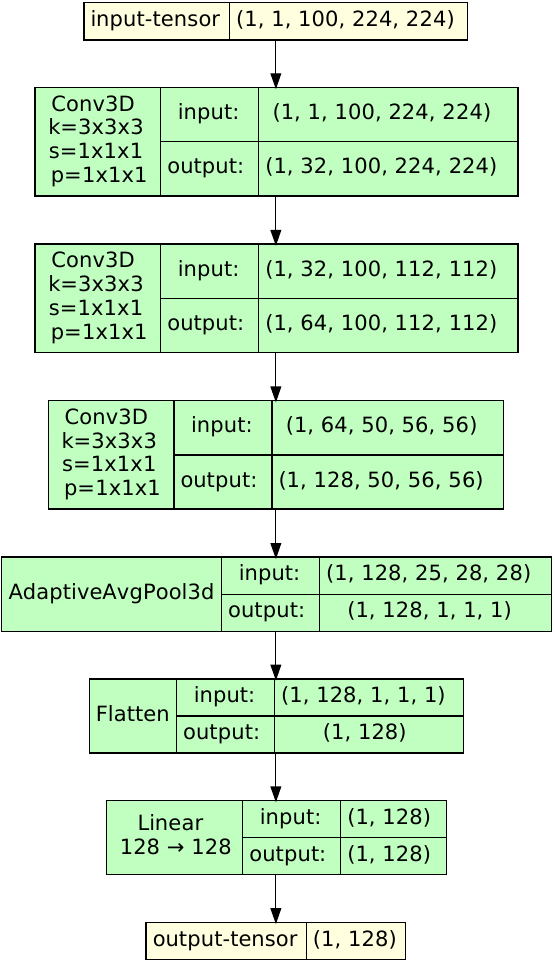}  

  \caption{Face segments.}
  \label{fig:face-segments}
\end{subfigure}\\
\caption{Architectures of the specialized encoders, namely rPPG, audio-visual emotion, head and gaze movement, and face segmentation encoders.}
\label{fig_architectures}
\end{figure*}

\subsection{Faliure Cases}

In Figure~\ref{fig:qualitative_results_bad}, we present failure cases of DF-MoE to point out some of its gaps. 
For the real video on the left-hand side, the AV Transformer and AVFF features are mainly responsible for the incorrect classification. The right-hand side example shows a fake video misclassified as real, where the model disproportionately attributes too much importance to emotion features. Our inspection of these predictions reveals persistent, inaccurate values of \emph{happiness} and \emph{fear}, that are misaligned with the actual emotions present in the audio-video streams. Together, these edge cases highlight two potential directions of improvement for our approach. First, we need to explicitly maintain a contribution balance when a small number of features begin to overshadow the rest of the cues. Second, we need to manage the sensitivity of the final classification to the precision of the upstream frozen feature extractors. 

\subsection{Implementation Details}

In this section, we provide details about the specialized encoders employed in our pipeline. In Figure~\ref{fig_architectures}, we illustrate the architectures that are briefly presented in the main paper for the rPPG, face segmentation, head movement and emotion encoders.

\end{document}